\documentclass{article}
\usepackage{arxiv}

\usepackage{lineno,hyperref}
\modulolinenumbers[5]

\usepackage{graphicx,natbib,amssymb}
\usepackage{algorithm}
\usepackage[noend]{algpseudocode}
\usepackage{tabularx}
\usepackage{hhline}

\usepackage{subfigure}
\usepackage{latexsym}
\usepackage{amsmath}
\usepackage{multirow}
\usepackage{multicol}
\usepackage{url}
\usepackage{booktabs}

\usepackage{header}

\title{Multi-Sense embeddings through a word sense disambiguation process}

\author{
  Terry Ruas\thanks{The final, published version of this article is available online in the Expert Systems with Applications Journal. Please check the final publication record for the latest revisions to this article. \url{https://doi.org/10.1016/j.eswa.2019.06.026}} \\
  Department of Computer and Information Science\\
  University of Michigan - Dearborn\\
  4901 Evergreen Rd, Dearborn, MI 48128, USA\\
  \texttt{ruas@uni-wuppertal.de} \\
  \texttt{truas@umich.edu} \\
   \And
William Grosky \\
Department of Computer and Information Science\\
  University of Michigan - Dearborn\\
  4901 Evergreen Rd, Dearborn, MI 48128, USA\\
  \texttt{wgrosky@umich.edu} \\
  
  \AND
  Akiko Aizawa \\
  National Institute of Informatics\\
  100-0003 Tokyo, Japan\\
  \texttt{aizawa@nii.ac.jp} \\
}

\begin{document}
\maketitle

\thispagestyle{firststyle}


% \author{Terry Ruas*, William Grosky}
% \ead{ruas@uni-wuppertal.de, truas@umich.edu, wgrosky@umich.edu}
% \address{University of Michigan - Dearborn, 4901 Evergreen Rd, Dearborn, MI 48128, USA}
% %\fntext[myfootnote]{Since 1880.}

% \author{Akiko Aizawa}
% \address{National Institute of Informatics, 100-0003 Tokyo, Japan}
% \ead{aizawa@nii.ac.jp}

%\author[mymainaddress]{\corref{mycorrespondingauthor}}

%\author[mysecondaryaddress]{National Institute of Informatics}
%\cortext[mycorrespondingauthor]{Corresponding Author}
%\ead{truas@umich.edu, wgrosky@umich.edu, aizawa@nii.ac.jp}

%\address[mymainaddress]{4901 Evergreen Rd, Dearborn, MI 48128, USA}
%\address[mysecondaryaddress]{〒100-0003 Tokyo, Japan}

\begin{abstract} %#R2 - 2
 Natural Language Understanding has seen an increasing number of publications in the last few years, especially after robust word embeddings models became prominent, when they proved themselves able to capture and represent semantic relationships from massive amounts of data. Nevertheless, traditional models often fall short in intrinsic issues of linguistics, such as polysemy and homonymy. Any expert system that makes use of natural language in its core, can be affected by a weak semantic representation of text, resulting in inaccurate outcomes based on poor decisions. To mitigate such issues, we propose a novel approach called \emph{Most Suitable Sense Annotation (MSSA)}, that disambiguates and annotates each word by its specific sense, considering the semantic effects of its context. Our approach brings three main contributions to the semantic representation scenario: (i) an unsupervised technique that disambiguates and annotates words by their senses, (ii) a multi-sense embeddings model that can be extended to any traditional word embeddings algorithm, and (iii) a recurrent methodology that allows our models to be re-used and their representations refined. We test our approach on six different benchmarks for the word similarity task, showing that our approach can produce state-of-the-art results and outperforms several more complex state-of-the-art systems.
 
\end{abstract}

%Keywords Multi-sense embeddings\sep natural language processing \sep word similarity\sep synset
%\MSC[2010] 00-01\sep  99-00

%%PRE-PRINT LINK

%\linenumbers

\section{Introduction}\label{sec:intro}
Semantic analysis is arguably one of the oldest challenges in Natural Language Processing (NLP) and still present in almost all its downstream applications. Even among humans, the precise definition of semantics cannot be agreed upon, which leads to multiple interpretations of text, making computational semantics even more challenging~\citep{Putnam:70}.

%Computational semantics is not an easy task.

Despite being a classical problem, the popularity of semantic analysis continues to draw the attention of numerous research projects in many different areas of study, under the rubric of \emph{semantic computing}. For example, ~\cite{Grosky:17} analyzed 2,872 multimedia publications (e.g. papers, journals, reports) between 2005 and 2015, revealing an increasing trend in publications involving \emph{Semantics} and \emph{Contextual Aspects} in different areas of multimedia. In these publications, methods applying different techniques try to capture semantic characteristics of text documents using such state-of-the-art approaches as latent semantic analysis, word embeddings, machine learning, and artificial neural networks.

After recent contributions~\citep{Mikolov_a:13,Mikolov_b:13,Penni:14}, word embeddings techniques have received much attention in the NLP community. These approaches represent words or phrases by real vectors that can be used to extract relationships between them. The overall performance of these algorithms have demonstrated superior results in many different NLP tasks, such as chunking~\citep{Dhillon:11}, meaning representation~\citep{Bordes:12}, machine translation~\citep{Mikolov_b:13}, relation similarity~\citep{Iacobacci:15,Mikolov_c:13}, sentiment analysis~\citep{Socher:13}, word sense disambiguation (WSD) ~\citep{Camachob:15,Chen:14,Navigli:09}, word similarity~\citep{Chen:15,Iacobacci:15,Neela:14} and topic categorization~\citep{Taher:17}.

Notwithstanding their robustness, however, most conservative word embeddings approaches fail to deal with polysemy and homonymy problems~\citep{Li:15}. Recently, researchers have been trying to improve their semantic representations by producing multiple vectors (multi-sense embeddings) based on a word's sense, context, and distribution in the corpus~\citep{Huang:12,Reisinger:10}. Another concern with traditional techniques is that they often neglect exploring lexical structures with valuable semantic relations, such as WordNet (WN)~\citep{Fellbaum:98}, ConceptNet~\citep{LiuCN:04} and BabelNet~\citep{Navigli:12}. Some publications take advantage of these structures and use them to develop multi-sense representations, improving the overall performance of their techniques ~\citep{Iacobacci:15,Iacobacci:16,Li:15,Mancini:17,Taher:16,Rothe:15}.

In this paper, we propose a model that allows us to obtain specific word-sense vectors from any non-annotated text document as input. For this, we extend the disambiguation algorithm presented by~\cite{Ruas_b:17} to find the most suitable sense of a word based on its context, which is later trained into a neural network model to produce specific vectors. The benefits of our approach are five-fold. First, we provide an unsupervised annotation algorithm that takes the context of each word into consideration. Second, our model disambiguates words from any part-of-speech (POS), mitigating issues with polysemy and homonymy. Third, the annotation and training steps in our approach are independent, so if more robust algorithms are available they can be easily incorporated. Fourth, the generated word-sense vectors keep the same algebraic properties as traditional word embedding models, such as $\emph{vec(king) - vec(man) + vec(woman) $\eqsim$  vec(queen)}$. Lastly, our architecture allows the produced embeddings to be used in the system recurrently, improving the quality of its representation and the disambiguation steps. To validate the quality of our work, we test our approach on 6 different benchmarks for the word similarity task, showing that our models can produce good results and sometimes outperform more complex current state-of-the-art systems.

The remainder of this paper is organized as follows. Section~\ref{sec:relwor} introduces the related work in word embeddings and multi-sense embeddings. In Section~\ref{sec:alg}, our proposed technique to annotate, disambiguate, and embed word-senses is described in detail. Section~\ref{sec:mse} explains the theory behind the various metrics for the multi-sense embeddings used in this paper. In Section~\ref{sec:expe}, we describe our experimental results and discuss the evaluation of the proposed algorithms in comparison with those of different systems. Lastly, in Sections~\ref{sec:limitations} and~\ref{sec:concl} we explain some of the strengths and weaknesses of our approaches, and present a few final considerations about our work, including its future directions.

%%%%%%%%%%%%%%%%%%%%%%%%%%%%%%%%%%%%%%%%%%%%%%%%%%%%%%%%%%%%%%%%%%%%%%%%%%%%%%%%%%%%%%%%%%%%%%%%%%%%%%%%%

\section{Related work}\label{sec:relwor}
%general text explaining the divisions?

%single sense embeddings
The distributed representation of words from documents has received substantial attention from the NLP community in the last few years, especially after the extremely popular \emph{word2vec} approach was proposed by~\cite{Mikolov_a:13,Mikolov_b:13}. However, the idea that words with similar contexts should have similar meanings goes back to the 20th century, with the \emph{Distributional Hypothesis}~\citep{Harris:54}. Later, the presence of these words would be described by count-based methods as in bag-of-words (BOW)~\citep{Salton:75}. Due to its simplistic methodology, however, the BOW approach has some drawbacks, such as data sparsity, loss of word order, and high dimensionality, to name a few.~\cite{Bengio:03} tries to solve the latter problem (dimensionality) by proposing a neural probabilistic language model that learns a representation while keeping a compact probability distribution of word sequences.~\cite{Collobert:08} later defined a faster general single convolutional neural network (CNN) architecture showing that multitask learning and semi-supervised learning can improve the generalization in shared tasks (e.g. POS tagging, morphological segmentation, named entity recognition, word similarity). Besides these, other language prediction models are also popular in the NLP community~\citep{Bordes:12,Turian:10,Turney:10,Zou:13}.

It is undeniable that word2vec's contributions with continuous skip-gram (SG) and continuous bag-of-words (CBOW) from~\cite{Mikolov_a:13,Mikolov_b:13} brought a legion of new publications to NLP, or more specifically, to the word embeddings arena. Its popularity is due to, among other things, its efficient log-linear neural network language model and its low-dimensionality vector representation. Both approaches produce vector representations of words, and those with similar contexts tend to have similar values. This theory was first described in the \emph{Distributional Hypothesis}~\citep{Harris:54} and popularized by~\cite{Firth:57}, which defended the concept of ``\emph{a word is characterized by the company it keeps}''. In the CBOW training model, one tries to predict a word given its neighboring context, while skip-gram does the inverse, predicting the context given a target word. Additional word embedding representations are also explored by SENNA~\citep{Collobert:11}, GloVe~\citep{Penni:14}, and fastText~\citep{Bojanowski:17}. Our approach builds on word2vec's predictive model to train a disambiguated corpus into specific word-sense vectors that can be applied to several NLP tasks. This will allow us to deal with one of the most important problems in traditional word embeddings techniques, the one vector representation per word property. %#R2-4  

%WSD part
Encouraged by the robustness of word embeddings,~\cite{Oele:18} combine word-sense, context and word-definition embeddings to support their disambiguation system. They extend the Lesk algorithm~\citep{Lesk:86} in two different scenarios using AutoExtend~\citep{Rothe:15} as their embedding training algorithm, the first using a distributional thesauri, and the second using WordNet hierarchical structure. In~\citep{Pelevina:16}, they propose a multi-stage system that learns single vector word representations, calculates word similarity graphs, infers word-senses using ego-network clustering, and aggregates word vectors with their possible word-sense vectors. In contrast with these approaches, we use only single vector word embeddings to support our disambiguation process a single time. Once our annotated corpus is trained, using a traditional word embeddings implementation, we can perform our disambiguation step by considering the specific word-sense embeddings directly. In addition, we do not rely on any extra parameters, other than those required by word2vec. The WSD technique proposed in this paper is inspired by the approach of~\cite{Ruas_b:17}, which produces word-sense representations for a given word based on its context. Even though disambiguation is a crucial component in our approach, the presented experiments and discussions focus more on how the combination of WSD and word embeddings can be mutual beneficial in the word similarity task~\citep{Iacobacci:16}. We do have future plans to compare our WSD technique with alternative methods and see how this affects the overall experimental results in this paper, but for now this is beyond our scope. For more details of the WSD field, we suggest~\citep{Navigli:09}'s survey, in which supervised, unsupervised, and knowledge-based approaches are discussed in depth. %#R2-4,  R2-5 - In contrast [...]

%multi sense embeddings
Nearly all publications in single vector word embeddings suffer from the same problem, in that words having multiple senses are represented by a unique vector. That is to say, polysemy and homonymy are not handled properly. For example, in the sentence ``This \emph{club} is great!'' it is not clear if the term \emph{club} is related to the sense of \emph{baseball club}, \emph{clubhouse}, \emph{golf club}, or any other appropriate sense. Systems that use standard word embeddings, such as word2vec or GloVe, will most likely represent all possible meanings for the term \emph{club} in one single-dimensional vector.

Some researchers try to solve this representation limitation by producing separate vectors for each word-sense. Even though the number of publications in this area is still small, their early findings demonstrate encouraging results in many NLP challenges~\citep{Li:15}. One of the earliest models was proposed by~\citep{Huang:12,Reisinger:10}. Both of these work with the concept of clustering word-senses by their context. \cite{Huang:12} introduce a neural network language model capable of distinguishing the semantics of words by considering their global (entire document) and local (surrounding words) contexts. In \citep{Reisinger:10}, they follow a probabilistic approach to produce a multi-prototype vector space model, using word-sense discovery to evaluate a word's context. They set a number, $K$, of clusters to represent the different contexts where the word is used. We, on the other had, combine the prior knowledge of WordNet and word embeddings to extract the many meanings of a word in an unsupervised manner. Since we produce a vector representation for each word-sense, the global meaning of a word in a document is the combination (average) of all senses for that word. This way, we do not need to rely on cluster parameters, which would increase the complexity of our approach. ~\cite{Trask:15} extend ~\cite{Huang:12}'s model by leveraging supervised NLP labels, instead of relying on unsupervised clustering techniques to produce specific word-sense vectors. We follow a similar idea and let the words define their own senses according to the context where they are located. However, our approach also takes advantage of the lexical structure provided by WordNet~\citep{Fellbaum:98}, which helps us to identify the implicit relationships between the words. In addition, our model's results can be fed into a word embeddings technique and re-used to improve itself recurrently, with respect to the disambiguation and embeddings steps (Section~\ref{ssec:mssa-nr}).  
%#R2-4,  R2-5

Other techniques also explore probabilistic models to learn their own representation for each sense.~\cite{Tian:14} design an efficient expectation maximization algorithm integrated with the skip-gram model to avoid the issues brought by clustering-based approaches. Another modification of skip-gram is proposed by~\cite{Neela:14}, in which they introduce the \emph{Multi-Sense Skip-Gram} (MSSG) model. Their technique performs word-sense discrimination and embedding simultaneously, improving its training time. In the MSSG version, they assume a specific number of senses for each word, while in the \emph{Non-Parametric Multi-Sense Skip-Gram} (NP-MSSG) this number varies, depending on the word. As in the NP-MSSG model, our approach also does not limit the number of word-senses for each word, but we used the CBOW algorithm instead of the skip-gram training model to produce our word-sense embeddings. Since MSSG and NP-MSSG both perform the disambiguation and embeddings in a unique process in their algorithms, this prevents them from exploring the benefits of other word embeddings techniques. We, on the other hand, have a modular system, in which disambiguation and word embeddings are done separately. Thus, our approach can use other lexical databases for extracting the semantic relationships between words, or incorporate new kinds of word embeddings techniques. Additionally, in contrast with \citep{Neela:14}, all our algorithms take advantage of the WordNet semantic network to help our system to better identify the possible senses for each word.  %#R2-4,  R2-5

%multi-sense + lexicography datasets
In multi-sense embeddings approaches the use of lexical resources to improve their performance in NLP tasks is quite common. WordNet\footnote{\url{https://wordnet.princeton.edu}}, ConceptNet\footnote{\url{http://conceptnet.io}} and BabelNet\footnote{\url{https://babelnet.org}} are examples of popular choices to help obtain word-sense vectors. Based on BabelNet, the system of~\citep{Iacobacci:15} learns word-sense embeddings for word similarity and relation similarity tasks, moving from a word to a sense embedding representation. Our choice for WordNet is supported by its open source policy under any circumstances, which as in ConceptNet, is very attractive. Moreover, WordNet is fully integrated with the Natural Language Toolkit (NLTK) in Python, which is heavily used in our implementation, making its choice preferable over other lexicographic resources, for now. ~\citep{Rothe:15,Rothe:17} also use WordNet in their \emph{AutoExtend} to produce token embeddings from a set of synonyms (\emph{synsets}) and lexemes, using a pre-existing word embeddings model. Similar to our model, their approach is independent of any word-type representation, so it can be easily translated to other learning techniques. They assume that a word can be represented by the sum of its lexemes, so their methodology puts words, lexemes, and synsets in the same vector space. As in~\citep{Rothe:15}, we explore the use of synsets as well, but beyond that we take into account their respective \emph{glosses}\footnote{\url{https://wordnet.princeton.edu/documentation/wngloss7wn}}, which is not considered in their model and aggregates solid information to the disambiguation process. As a result, our semantic representation obtains better results when the context is available in the word similarity task (Section~\ref{ssec:cws}). In the iterative version of MSSA (Section`\ref{ssec:mssa-nr}), we can also use our own produced vectors to improve the disambiguation step. Additionally, our annotated corpus is self-contained with respect to its representation in the lexical database. In other words, from the annotation results or the word embeddings model keys, one can retrieve the lexeme, synset, word or any other information available in WordNet for a specific word (Section~\ref{ssec:s2v}).  %#R2-4, R2-5

Continuing with the multi-sense embeddings approaches,~\cite{Camacho:16} propose a powerful semantic vector representation called \emph{NASARI} which is extended to a multilingual scenario. Their system uses BabelNet as a lexical database and provides vector representations of concepts in several different languages over a unified semantic space. This approach, combined with all lexicons incorporated by BabelNet, gives NASARI a robust architecture. As in in our approach, SensEmbed~\citep{Iacobacci:15} produces word-sense vectors based on a disambiguated and annotated corpus. However, their disambiguation process relies on Babelfy~\citep{Moro:14}, which combines WSD and entity linking to build a dense graph representation of sense candidates for each word, using BabelNet as a backbone. In order to take advantage of BabelNet's semantic structure, they need to consider a parameter called \emph{graph vicinity factor} in their overall system. We, on the other hand, do not require any parameter tuning during the disambiguation step to explore WordNet's structure. More recently, the algorithm of ~\cite{Mancini:17} associates words to the most connected senses in a sentence to produce their embeddings. In their approach, sentences are parsed and only word-senses with more than a certain number of connections (cut-off) with other words in the same sentence are selected. These connections illustrate the relationships (edges) between the nodes (synsets) in BabelNet. In the end, both word and senses embeddings are represented in the same vector space. Differing from all these approaches, we only rely on the training corpus available for the disambiguation step in combination with WordNet. No other multi-lexicographic databases are used, nor extra hyperameters considered to incorporate external information about the word-sense. In addition, our techniques (Sections~\ref{ssec:mssa} and \ref{ssec:dmssa}) produce word-sense embeddings that can also be used to improve the WSD process in a recurrent manner (Section~\ref{ssec:mssa-nr}). %R2-4,  R2-5

The system of~\cite{Chen:14} performs WSD on the words and uses them to learn word-sense representations from the relevant occurrences through two approaches: L2R (left to right) and S2C (simple to complex). In L2R, they disambiguate words from left to right, whereas S2C selects only word-senses that reach a certain similarity threshold to represent a word. We believe that considering just one order in the disambiguation step or only specific word-senses leads to a poor and biased representation of semantics. In MSSA, we explore the contextual effects of all senses available for a word and its neighbors in a bi-directional fashion. In other words, our techniques use a bi-directional approach, since given a target word it considers the semantic influence of its previous (left to right) and successive (right to left) neighboring words. This prevents us from ignoring possible candidates even if they are not that frequent in the text, something that is not explored by \cite{Chen:14} %#R2-4, R2-5

\cite{Chen:15} use WordNet's glosses to produce word-sense vectors via a convolutional neural network, which are used as input into an MSSG variant. Their system uses the sentence in the glosses as positive training data and replaces random words (controlled by a parameter) to create negative training data, in order to minimize a ranking loss objective. Our techniques tackle this situation using two different approaches that are more effective: average the vectors from the glosses for each synset (Section~\ref{ssec:mssa}) and re-use the produced synset embeddings to support the disambiguation task iteratively  (Section~\ref{ssec:mssa-nr}). Both options provide a simpler alternative and perform better in the word similarity task (Section~\ref{sec:expe}). In addition, there is no extra training or hyperparameter adjustments other than those required by the standard word2vec implementation. Therefore, with less training phases and fewer hyperparameter adjustments our techniques are able to perform better for all datasets compared for at least one of the metrics used in the word similarity task (Section~\ref{sec:mse})    %#R2-4,  R2-5

%%%%%%%%%%%%%%%%%%%%%%%%%%%%%%%%%%%%%%%%%%%%%%%%%%%%%%%%%%%%%%%%%%%%%%%%%%%%%%%%%%%%%%%%%%%%%%%%%%%%%%%%%%%%%%%%%%%%%%%%%%

\section{Synset disambiguation, annotation, and embedding} \label{sec:alg}
\begin{figure}[!htb]
\centering
\includegraphics[width=\textwidth]{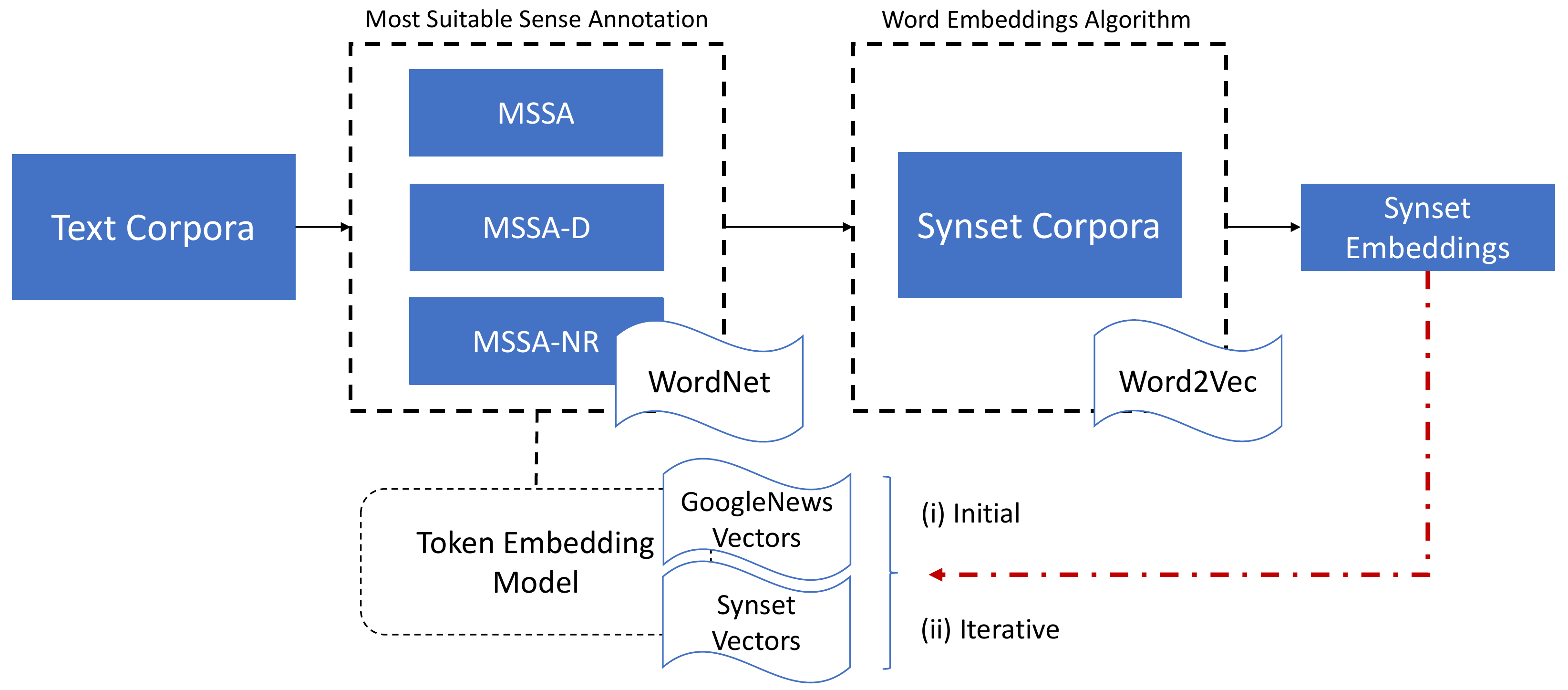}
\caption{General system architecture of MSSA, MSSA-D and MSSA-NR.}\label{fig:Sysarch}
\end{figure}

The main idea of our process is to have a modular system with two independent tasks: (i) disambiguation followed by annotation, and (ii) token embeddings training. This configuration allows us to incorporate more robust techniques in the future, especially for the training step. The disambiguation and annotation module require nothing more than a lexical database, English corpora, and a token embeddings model to transform word-based documents into word-sense-based documents. As for the embeddings training module, any word embeddings algorithm that can represent tokens in a vector space is suitable. In the following sections we will explain the details of our approach illustrated in Figure~\ref{fig:Sysarch}.

In the first task, we process a collection of articles (documents) from two Wikipedia Dumps (Section~\ref{ssec:corpus}) to transform each word in the corpus into a synset by using WordNet as our lexical resource~\citep{Fellbaum:98,Miller:95}. This is done through one of the proposed algorithms: \emph{Most Suitable Sense Annotation (MSSA)} (Section~\ref{ssec:mssa}), \emph{Most Suitable Sense Annotation N Refined (MSSA-NR)} (Section~\ref{ssec:mssa-nr}) and \emph{Most Suitable Sense Annotation - Dijkstra (MSSA-D)} (Section~\ref{ssec:dmssa}). In the second task, we use word2vec~\citep{Mikolov_a:13,Mikolov_b:13} to train this synset corpus and obtain $n$-dimensional vectors of each word-sense (multi-sense embeddings).

% #R1-3 - Last two sentences
In their initial form, both MSSA and MSSA-D use Google News vectors\footnote{\url{https://code.google.com/archive/p/word2vec/}} to help disambiguate the word-senses in the corpus. MSSA works locally, trying to choose the best representation for a word, given its context window. MSSA-D on the other hand, has a more global perspective, since it considers the most similar word-senses from the first to the last word in a document. For the word embeddings training module. Once the synset embeddings models are available, we can feed the system again, using the output vectors from the previous pass, and improve the disambiguation step in either the MSSA or the MSSA-D algorithms, relieving them from the original Google News vectors' dependency. We call this approach MSSA-NR, where $N$ is the number of feedback iterations used. This recurrent characteristic was not explored by any of the related works (Section~\ref{sec:relwor}) nor the compared systems in the experiments (Section~\ref{sec:expe}). Different from other systems~\citep{Chen:14,Chen:15,Rothe:15}, our method has only one training phase and does not rely on any extra hyperparameters, other than those required in the original word2vec implementation. In addition, since all proposed MSSA approaches are performed in the raw text directly and prior to any training model, they can be easily incorporated into any NLP pipeline, independently of the task. In other words, MSSA would work the same way as any common pre-processing activity (e.g stemming, stop-words removal, lower case). % #R1-3/#R2-3

%Fix this for two column alignment in new template

\subsection{Most Suitable Sense Annotation (MSSA)}\label{ssec:mssa}
As \cite{Ruas_b:17} present, each evaluated word $w_{i}$ takes into consideration its context, represented by its surrounding neighboring words, $w_{i-1}$ and $w_{i+1}$, as illustrated in Algorithm~\ref{alg:mssa}. We also use WordNet as our lexical database to extract all synsets from each word in the text, but unlike \citep{Ruas_b:17}, our algorithm works for any word mapped in WordNet, not just for nouns. In our approach, all text is first preprocessed, normalizing all tokens in lowercase, removing punctuation, html tags, numbers, common English stopwords, and discarding all words not present in WordNet. The list of common stopwords used is obtained directly through the Natural Language Toolkit (NLTK) library in Python. After this initial data cleaning, we extract all pairs of synsets and glosses for each word $w_{i}$ in a sliding context window of 3 words. (lines~\ref{mssa:sgstart}:\ref{mssa:sgend}). Our context sliding window is similar to the one used in CBOW, proposed by \cite{Mikolov_a:13}, which uses the context to predict a given word. However, since our algorithm considers all synsets from $w_{i}$, $w_{i-1}$ and $w_{i+1}$, we currently limit this word context window to restrict the necessary number of comparisons, so as to infer the most suitable meaning for $w_{i}$. It is in our plans to incorporate a larger context without compromising the overall performance for this step. Next, after removing common English words from the glosses, we retrieve and average the embeddings from the remaining tokens in each gloss, which we call \emph{gloss-average-vector}, by using Google News vectors\footnote{\url{https://code.google.com/archive/p/word2vec/}}. If there are no remaining tokens in the gloss or no vectors in the model, an empty vector will be assigned for that synset-gloss pair. However, this scenario is very unlikely, since the words in the glosses have their vector extracted from a model trained on a huge corpus.  This process is done for all synset-glosses for each element $s_{c}$ ($current\_candidates$), $s_{f}$ ($former\_candidates$) and $s_{l}$ ($latter\_candidates$) (lines~\ref{mssa:gavgstart}:\ref{mssa:gavgend}), where $M$, $N$ and $P$ represent the total number of available synset-glosses per synset, respectively. After the gloss-average-vectors for each synset in a particular position of the sliding window are obtained, we calculate the cosine similarity of all synsets of the $current\_candidates$ against those of the $former\_candidates$ and the $latter\_candidates$, returning the synset for $current$ (in each case) with the highest score, as lines~\ref{mssa:cosine-f} and~\ref{mssa:cosine-l} describe. Finally, we add the synset with the highest value to our list of tokens, in order to represent this occurrence of $w_{i}$ in our new synset corpus (line~\ref{mssa:mssa}). It is important to mention that the first and the last words in our corpus are treated differently, since they do not have a complete context window available, as in word2vec (lines~\ref{mssa:first} and \ref{mssa:last}).

\begin{algorithm}[H]
\small
   \caption{Most Suitable Sense Annotation (MSSA)}
   \label{alg:mssa}
    \begin{algorithmic}[1]
    \Require $d = \{w_{i},...,w_{n}\} : w_{i} \in$ \emph{lexical database} (WordNet) \label{mssa:req1}
      \Function{MSSA}{$d$, $tm$, $ld$}\Comment{for $d$ - document containing words $w_{n}$, $tm$ - trained word embedding model, $ld$ - lexical data base}
        \State \emph{list\_of\_tokens} $=$ $\emptyset$
        \For{$i=0$ to $n$} \label{mssa:sgstart}
            \State $current =$ synset-glosses($w_{i}$, $ld$)
            \If {$i \neq 0 \land i \neq n$}
                \State $former =$ synset-glosses($w_{i-1}$, $ld$)
                \State $latter =$ synset-glosses($w_{i+1}$, $ld$)
            \ElsIf {$i = 0$} \label{mssa:first}
                \State $former =$ $\emptyset$
                \State $latter =$ synset-glosses($w_{i+1}$, $ld$)
            \Else \label{mssa:last}
                \State $former =$ synset-glosses($w_{i-1}$, $ld$)
                \State $latter =$ $\emptyset$
            \EndIf \label{mssa:sgend}
            \State $current\_candidates =$ $\emptyset$, $former\_candidates =$ $\emptyset$ and \label{alg:complexity-start} \\ $latter\_candidates =$ $\emptyset$ 
            \For{$s_{c} \in \{\emph{current}\}$, $s_{f} \in \{\emph{former}\}$ and $s_{l} \in \{\emph{latter}\}$} \Comment{for $0\leq c\leq M$, $0\leq f\leq N$ and $0\leq l\leq P$}\label{mssa:gavgstart}
                \State \textbf{Add} gloss-avg-vec($s_{c}$, $tm$) \textbf{to} $current\_candidates$
                \State \textbf{Add} gloss-avg-vec($s_{f}$, $tm$) \textbf{to} $former\_candidates$
                \State \textbf{Add} gloss-avg-vec($s_{l}$, $tm$) \textbf{to} $latter\_candidates$   \label{mssa:gavgend}
            \EndFor
            \State $u =$ $argmax_{s_{c1}}\{$cosine-similarity(\emph{current\_candidates}$, $\emph{former\_candidates}$)\}$ \label{mssa:cosine-f}
            \State $w =$ $argmax_{s_{c2}}\{$cosine-similarity(\emph{current\_candidates}$, $\emph{latter\_candidates}$)\}$  \label{mssa:cosine-l}
            \State \textbf{Add} the synset ($s_{c1}$ or $s_{c2}$) with the highest produced cosine similarity \textbf{to} $\emph{list\_of\_tokens}$   \label{mssa:mssa}
        \EndFor
        \State \textbf{return} $list\_of\_tokens$
       \EndFunction

\end{algorithmic}
\end{algorithm}

In the initial configuration, we use Google News vectors as our standard word embeddings model ($tm$ in lines~\ref{mssa:gavgstart}:\ref{mssa:gavgend}), which was trained over 100 billion words and contains 300-dimensional vectors for 3 million unique words and phrases~\citep{Mikolov_b:13}. This approach can also work recurrently, using the current synset embeddings to be fed back into our system, so that in the next iteration we use our previously calculated vectors of synsets to disambiguate the word-senses in the corpus. In this modality, it is not necessary to calculate the gloss-average-vector for each synset-gloss again, since we can use the synset embeddings directly to disambiguate our training corpus.

\subsection{Most Suitable Sense Annotation N Refined (MSSA-NR)}\label{ssec:mssa-nr}
As mentioned above, once we have trained our model based on synset tokens, we can use these output synsets vectors directly for another pass of our algorithm, bypassing the gloss-average-vector calculation. As we do not need to calculate the gloss-average-vector for each recurrence after the first one, each future pass will take less time than the first pass.  We hypothesize that by using disambiguated and granular embeddings we will obtain a more refined synset model. The algorithm for this approach is similar to the one presented in Section~\ref{ssec:mssa}, so we are still using the same cleaned training corpus composed of Wikipedia articles, but some of the steps are slightly different.

We identify this refined approach as \emph{MSSA-NR}, where $N$ represents the number of feedback iterations used. Algorithm \ref{alg:mssa-nr} starts in a similar fashion to Algorithm \ref{alg:mssa}, as we also use WordNet as our lexical database and still work with the same sliding context window for the words. The main difference occurs between lines \ref{mssa-nr:systart} and \ref{mssa-nr:sysend}, where, since our embeddings consist of synsets, we do not need to extract the pairs of synset-glosses and calculate the gloss-average-vector for each synset. Instead, we just extract all synsets available in WordNet for $w_{i}$ ($current$), $w_{i-1}$ ($former$), $w_{i+1}$ ($latter$) and directly retrieve their respective vector embeddings from the synset model trained (lines~\ref{mssa-nr:systart}:\ref{mssa-nr:sysend}), where $Q$, $R$ and $S$ represent their total number of available synsets per word. Since MSSA-NR is using an embedding model on the same corpus on which it was first generated, all the words will have at least one synset mapped, so there is no risk of not finding a vector for a given word-sense. After we retrieve the vector values for all synsets in the sliding window, we calculate the similarity of $current\_candidates$ against $former\_candidates$ and $latter\_candidate$, returning the synsets for $current\_candidates$ with the highest value in each case (lines~\ref{mssa-nr:cosine-f} and ~\ref{mssa-nr:cosine-l}). As in MSSA, we also select the synset with the highest score to represent $w_{i}$ in our new synset corpus (line~\ref{mssa-nr:mssa-nr})

%#R1-3 - Mention the complexity in terms of big-o notation
Because we are using the word-sense embeddings from our previous pass, senses that never were selected to represent any word in the original corpus will not have a vector representation in our model. As a consequence, in the next iteration, these word-senses do not have to be verified, since they were not embedded in the first place. The hope is that, over many passes, the non-used word-senses are dropped and results will converge to some stable synset-value representation of our corpus. This will also contribute to a faster processing time, if compared to the MSSA approach, considering that the number of word-senses is reduced on each pass until it stabilizes. We can stop the process after a finite number of passes, when we are satisfied that the results do not change much, or when the cost incurred for running another pass of the algorithm is too high to justify another disambiguation and annotation round. More details about the overall complexity are provided in Section~\ref{ssec:complexity}.

\begin{algorithm}[H]
\small
   \caption{Most Suitable Sense Annotation N Refined (MSSA-NR)}
   \label{alg:mssa-nr}
    \begin{algorithmic}[1]
    \Require $d = \{w_{i},...,w_{n}\} : w_{i} \in$ \emph{lexical database} (WordNet) \label{mssa-nr:req1}
      \Function{MSSA-NR}{$d$, $tsm$, $ld$}\Comment{for $d$ - document containing words $w_{n}$, $tsm$ - trained synset embedding model, $ld$ - lexical data base}
        \State \emph{list\_of\_tokens} $=$ $\emptyset$
        \For{$i=0$ to $n$} \label{mssa-nr:systart}
            \State $current =$ synsets($w_{i}$, $ld$)
            \If {$i \neq 0 \land i \neq n$}
                \State $former =$ synsets($w_{i-1}$, $ld$)
                \State $latter =$ synsets($w_{i+1}$, $ld$)
            \ElsIf {$i = 0$} \label{mssa-nr:first}
                \State $former =$ $\emptyset$
                \State $latter =$ synsets($w_{i+1}$, $ld$)
            \Else \label{mssa-nr:last}
                \State $former =$ synsets($w_{i-1}$, $ld$)
                \State $latter =$ $\emptyset$
            \EndIf \label{mssa-nr:sysend}
            \State $current\_candidates =$ $\emptyset$, $former\_candidates =$ $\emptyset$ and \\ $latter\_candidates =$ $\emptyset$
            \For{$s_{c} \in \{\emph{current}\}$, $s_{f} \in \{\emph{former}\}$ and $s_{l} \in \{\emph{latter}\}$} \Comment{for $0\leq c\leq Q$, $0\leq f\leq R$ and $0\leq l\leq S$}\label{mssa-nr:vvstart}
                \State \textbf{Add} synset-vec($s_{c}$, $tsm$) \textbf{to} $current\_candidates$
                \State \textbf{Add} synset-vec($s_{f}$, $tsm$) \textbf{to} $former\_candidates$
                \State \textbf{Add} synset-vec($s_{l}$, $tsm$) \textbf{to} $latter\_candidates$  \label{mssa-nr:vvgend}
            \EndFor
            \State $u =$ $argmax_{s_{c1}}\{$cosine-similarity(\emph{current\_candidates}$, $\emph{former\_candidates}$)\}$ \label{mssa-nr:cosine-f}
            \State $w =$ $argmax_{s_{c2}}\{$cosine-similarity(\emph{current\_candidates}$, $\emph{latter\_candidates}$)\}$  \label{mssa-nr:cosine-l}
            \State \textbf{Add} the synset ($s_{c1}$ or $s_{c2}$) with the highest produced cosine similarity \textbf{to} $\emph{list\_of\_tokens}$   \label{mssa-nr:mssa-nr}
        \EndFor
        \State \textbf{return} $list\_of\_tokens$
       \EndFunction
\end{algorithmic}
\end{algorithm}

\subsection{Most Suitable Sense Annotation - Dijkstra (MSSA-D)}\label{ssec:dmssa}
\begin{figure}[htb]
\centering
\includegraphics[width=0.8\textwidth]{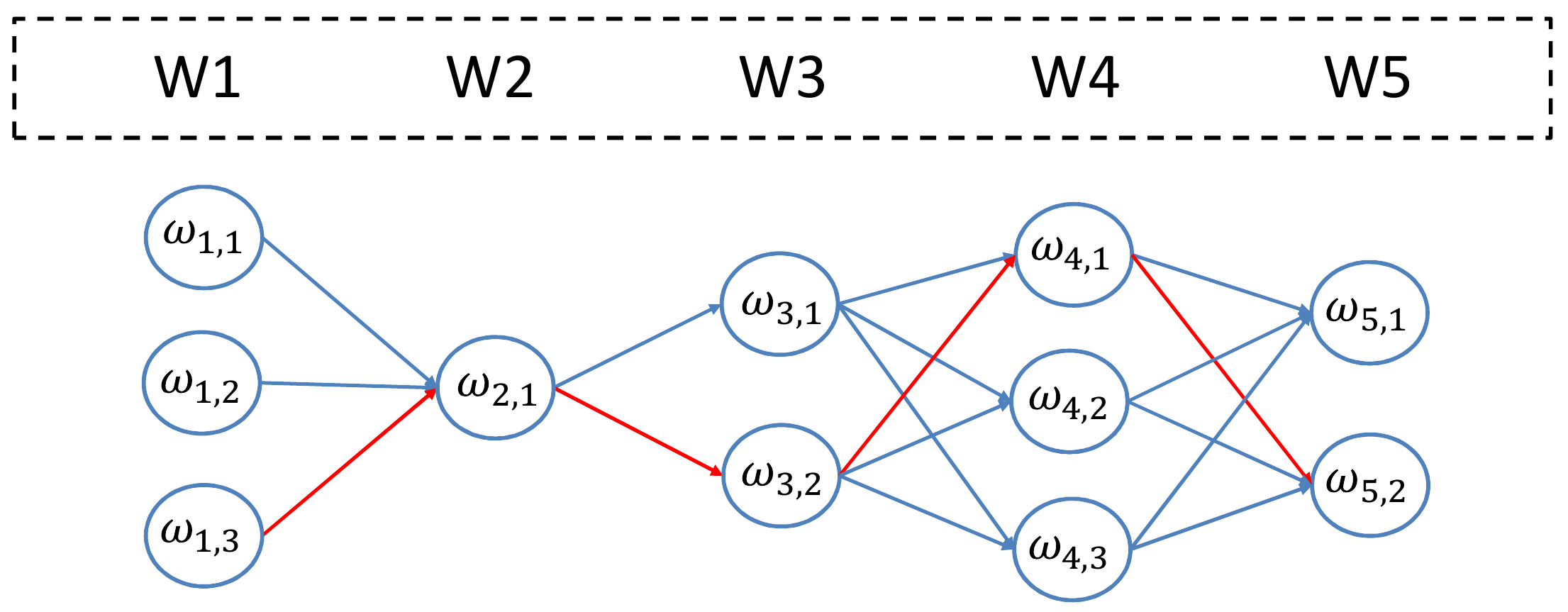}
\label{fig:Smssad}
\caption{MSSA-D illustration of the shortest path from $w_{1}$ to $w_{5}$ through their respective word-senses.}
\end{figure}

We also developed another variation for the MSSA algorithm, in which we model the documents in the corpus as a graph $Doc_{k}(N,E)$, where $Doc_{k}$ is the set of $k$ documents; $N$ is the set of nodes, represented by word-senses (synsets) and $E$ is the set of edges associating two nodes in document $k$. Inspired by Dijkstra's algorithm~\citep{Dijkstra:59}, we use a modified version of it to minimize the overall cost of moving from one node (synset) to another, for all the words in the document. The weights on the edges are the \emph{cosine distances} (1 - \emph{cosine similarity}) between the gloss-average-vector of two sequential word-senses. All the steps in the MSSA-D design are the same as the ones presented in Section~\ref{ssec:mssa} for MSSA (Algorithm~\ref{alg:mssa}), with the exception that there is no sliding context window for the disambiguation part. Different from MSSA, in MSSA-D we analyze the disambiguation problem globally, looking for the shortest distance from one word-sense to the next. Figure~\ref{fig:Smssad} illustrates a toy example of five words in which the highlighted path has the lowest cost, considering their word-senses $\omega_{n,m}$, where $n$ is the associated word position and $m$ its respective sense. In the end, the objective of this algorithm is the same as the ones presented in Sections~\ref{ssec:mssa} and~\ref{ssec:mssa-nr}, transforming a training corpus composed by words into a corpus of synsets to be trained by word2vec.

%Fix this for two column alignment in new template

As in MSSA, it is also possible to apply the MSSA-NR recurrent methodology into MSSA-D and reuse the produced synset embeddings instead of the Google News vectors. Considering this approach, there is, again, no need to calculate the gloss-average-vector for each word-sense. Instead, we can directly use the synset vectors available. In Section~\ref{sec:expe}, we describe the different setups used in our experiments to explore all of our techniques.
%As in Section~\ref{ssec:mssa}, we also consider the gloss-average-vector of every synsets to calculate the \emph{cosine distance} (1 - \emph{cosine similarity}) between them. The main idea in this algorithm is to create a sequence of word-senses, which will capture the overall semantics between words in a document with the lowest possible cost. This methodology is identified as \emph{MSSA-D} in our experiments.

\subsection{Synset to Embeddings (Synset2Vec)}\label{ssec:s2v}

After all words are properly processed into synsets, we use them as input in a word2vec implementation with CBOW as the training algorithm. This choice is justified, due to its popularity among the compared systems, with its reported superiority in performance, and its ability in capturing the context of words in large datasets~\citep{Bojanowski:17,Mikolov_a:13,Yao:17,Yu:14}.

%R2-3 Complexity big-o notation
In all MSSA variations, the goal is to transform a word-based document into a synset-based one. This will allow us to properly represent a word with multiple senses. Since the disambiguation step might consider all the senses of a word, its cost grows rapidly with the size of the documents and number of available senses per word. For a small to medium size training corpus, this is not a barrier, but for larger ones, such as the Wikipedia Dump used in our paper, this can demand a high amount of processing time. On the other hand, the number of tokens to be trained by the word embeddings algorithms is reduced, since only words that exist in WordNet are considered. In addition, the disambiguation process only needs to be executed once and it can be done in parallel for all MSSA techniques. Once the annotated synset training corpus is performed, one can use it in any desired activity, such as word similarity, document classification, text summarization, and train a word embeddings model.   %#R1-4

To keep our vectors interpretable - as pointed out in \citep{Pachenko:16} - across different platforms, we represent each word token as a key in the following format: $word\#synset\_offset\#pos$, where $word$ is the word itself, normalized in lowercase; $synset\_offset$ is an 8 digit, zero-filled decimal integer that corresponds to a unique word-sense, and $pos$ is a part-of-speech tag ($n$ for nouns, $v$ for verbs, $a$ for adjective, $s$ for adjective satellite and $r$ for adverb)\footnote{\url{https://wordnet.princeton.edu/documentation/wndb5wn}}. Since we have independent tasks for annotation and word embeddings training, if more robust techniques are proposed in the future, we can easily incorporate them.

\subsection{Complexity analysis}\label{ssec:complexity}
In this section we provide a detailed explanation on how to compute the average time complexity of each of our algorithms, MSSA, MSSA-NR, and MSSA-D.

For each of these algorithms, we have the same preprocessing step; namely, to associate a vector with each gloss in the current version of WordNet. However, this is done only once per WordNet version. In the current version of WordNet, assume that there are $S$ synsets, each having one gloss. For a synset $S_{i}$, let $g_{i}$ be the number of words in its gloss, after eliminating common English stop-words. Then the average number, $G$, of words/gloss is described as $ G = \frac{(g_{i}+\dots+g_{S})}{S}$. The vector associated with each word in a gloss is assumed to be the Google vector associated with that word and we assume that this vector can be found in $\theta(1)$ time, given the word. Given the Google vectors (pre-trained word embeddings model) for each word in a given gloss, the vector associated with this gloss is just the average of the vectors from the pre-trained model. Thus, for synset $S_{i}$, the time complexity of computing the vector associated with its gloss is $\theta(g_{i})$. We then have that the time complexity for computing the vectors associated with all glosses in WordNet is $\theta(g_{i}+\dots+g_{S}) = \theta(SG)$.

Considering MSSA, let us calculate the time complexity of computing the disambiguated synset corpus from the original one. Let $W$ be the number of words in the corpus. There are, on the average 2.89 synsets per word in WordNet\footnote{\url{https://wordnet.princeton.edu/documentation/21-wnstats7wn}}. Thus, $2.89W$ is a good approximation for the number of synsets processed by our algorithm. Given that each synset has one gloss vector and defining $gv_{i}$ as the number of gloss vectors processed for word $w_{i}$ of the corpus, we then get that $(gv_{1} +\dots+ gv_{W}) = 2.89W$. Based on lines~\ref{alg:complexity-start}:\ref{mssa:mssa} in Algorithm~\ref{alg:mssa}, the number of processed vectors is illustrated in Equation~\ref{eq:mssa_complexity1}.

\begin{equation}\label{eq:mssa_complexity1}
\small
    \begin{split}
        (gv_{1}gv_{2} + gv_{2}(gv_{1} + gv_{3}) + gv_{3}(gv_{2} + gv_{4}) +
        gv_{W-1}(gv_{W-2} + gv_{W}) + \\(gv_{W-1}gv_{W}) \le  (gv_{1} +\dots+ gv_{W})^{2} \le  2.89^{2}W^{2} = \theta(W^{2})
    \end{split}
\end{equation}

\noindent Thus, we compute $O(W^{2})$ cosine measures overall in the computation of the corresponding synset corpus. Assuming that $w2v$ is the running time of the standard approach to calculate the Google vectors using the CBOW approach, we have that the time complexity of MSSA is $O(W^{2}) + w2v$.

Now, let us consider MSSA-NR. By the argument above, it is seen that the first pass of MSSA-NR has time complexity $O(W^{2}) + w2v$. For each succeeding pass, we follow the same process as above, but with the synset vectors from the previous pass replacing the corresponding gloss vector. Thus, for $N$ passes, the time complexity is $O(NW^{2})+ N \ast w2v$.

Finally, for MSSA-D the underlying graph has $V$ vertices and $E$ edges. As shown above, $2.89W$ is a good approximation for the number synsets processed by our algorithm. Thus, $V = 2.89W$. Realizing that $gv_{i}$ is also equal to the number of synsets processed for word $w_{i}$ of the corpus, we have that the number of edges can be described as Equation~\ref{eq:mssa_complexity2} shows.

\begin{equation}\label{eq:mssa_complexity2}
\small
    \begin{split}
       E =  gv_{1}gv_{2} + gv_{2}gv_{3} + \dots gv_{W-1}gv_{W} + (gv_{W-1}gv_{W}) \\ \le  (gv_{1} +\dots+ gv_{W})^{2} \le  2.89^{2}W^{2}
    \end{split}
\end{equation}

Thus, since the time complexity of Dijkstra’s Algorithm is $O(V^{2})$, $O(V^{2} + E \log V)$ or $O(E + V\log V)$, depending on its implementation, we have that the time complexity of MSSA-D is $O(W^{2})$.

%%%%%%%%%%%%%%%%%%%%%%%%%%%%%%%%%%%%%%%%%%%%%%%%%%%%%%%%%%%%%%%%%%%%%%%%%%%%%%%%%%%%%%%%%%%%%%%%%%%%%%%%%%%%%%%%%%%%%%%%%%

\section{Multi-Sense embeddings measures} \label{sec:mse}
In a standard single $n$-dimensional vector representation (e.g. word2vec), there will be just one embedding for each token to calculate any similarity measure between them. In a multi-vector representation, each word is associated with a set of senses, each with a vector representation. Hence, the similarity of these word-senses need to be calculated in a different way. Both representations make use of several benchmarks to evaluate their performance in the word similarity task. These benchmarks can be grouped into two categories: with or without context information. In the first category, a similarity score is given for two words in isolation, without any extra information about them. In the second category, each word is presented with a sentence to help contextualize its semantic content.

Considering the multi-vector representation, two metrics were initially proposed \citep{Reisinger:10}: \emph{AvgSim} and \emph{MaxSim}. In AvgSim, word similarity is calculated by considering the average similarity of all word-senses for the pair, as shown in Equation~\ref{eq:avgsim}.

\begin{equation}\label{eq:avgsim}
\small
    \begin{split}
        AvgSim(u,w) = \frac{1}{NM} \sum_{i=1}^{N}\sum_{j=1}^{M} d(e(u,i),e(w,j))
    \end{split}
\end{equation}

where $u$ and $w$ are the words to be compared; $N$ and $M$ are the total number of available senses for $u$ and $v$, respectively; $d(e(u,i),e(w,j))$ is the similarity measure between the word-sense embeddings sets denoted by $e(u,i)$ and $e(w,j)$, between the $i^{th}$ sense of word $u$ and $j^{th}$ sense of word $w$. In MaxSim, the similarity is the maximum value among all pairs of word-sense embeddings, as illustrated in Equation~\ref{eq:maxsim}. In this paper, all similarity measures are calculated using the \emph{cosine similarity} between any two vectors.

\begin{equation}\label{eq:maxsim}
\small
    \begin{split}
        MaxSim(u,w) = \underset{1\leq i \leq N,1\leq j \leq M}{\max} d(e(u,i),e(w,j))
    \end{split}
\end{equation}

\cite{Reisinger:10} also propose \emph{AvgSimC} and \emph{MaxSimC}. These take into account the similarity of two words when their context is available. In this scenario, the context is represented by sentences in which the target words are used. For tasks with this setup, two words are evaluated with respect to their similarity and each of them will have a sentence illustrating their use. Both AvgSimC and MaxSimC are described by Equations~\ref{eq:avgsimc} and~\ref{eq:maxsimc} respectively. %if there's space talk about hard/soft cluster

\begin{equation}\label{eq:avgsimc}
\small
    \begin{split}
        AvgSimC(u,w) = \frac{1}{NM}\sum_{i=1}^{N}\sum_{j=1}^{M} P(u,c_{u},i) \\ P(w,c_{w},j)  \times d(e(u,i),e(w,j))
    \end{split}
\end{equation}

\begin{equation}\label{eq:maxsimc}
\small
    \begin{split}
        MaxSimC(u,w) = d(e_k(u,i),e_k(w,j))
    \end{split}
\end{equation}

where $P(u,c_{u},i) = d(e(u,i),c_{u})$, is defined as the similarity of the $i^{th}$ sense of word $u$ with its context $c_{u}$. The context ($c_{u}$) is obtained by averaging all vectors of the words in the sentence where $u$ is used. Different from single word vector representations, our model produces vectors for each word-sense, so when
we calculate the average vector of $c_{u}$, we need only to consider all available word-sense vectors. $e_k(u,i) = arg_{max} d(e(u,i),c_{u})$ is the maximum similarity obtained among all word-senses $e(u,i)$, with respect to its context $c_{u}$. All these terms are defined analogously for $w$ and $j$ as well. It is important to mention that the context defined for AvgSimC and MaxSimC are not related with the sliding context window presented in our approach (Section~\ref{sec:alg}).

\cite{Huang:12} argue that word representations should be discriminated by considering their surrounding words (\emph{local context}) and their role in the entire document (\emph{global context}). Their training model produces two vector types, one representing each word-sense and another for the word in the entire document, evaluated through \emph{LocalSim} and \emph{GlobalSim} respectively~\citep{Neela:14}. Unlike~\citep{Huang:12,Neela:14}, our approach does not produce global vectors during the training step, only local ones. Therefore, to obtain a global representation of a word, we average all word-sense vectors of $u$ and $w$ available to calculate their similarity, as Equation~\ref{eq:globalsim} shows.

\begin{equation}\label{eq:globalsim}
\small
    \begin{split}
        GlobalSim(u,w) = d(\check{\mu}(u,i),\check{\mu}(w,j))
    \end{split}
\end{equation}

where $\check{\mu}(u,i)$ and $\check{\mu}(w,j)$ represent the average of all word-sense vectors for $u$ and $w$. As for LocalSim, we can use the original MaxSimC instead, since they work under the same assumptions~\citep{Reisinger:10}.

%%%%%%%%%%%%%%%%%%%%%%%%%%%%%%%%%%%%%%%%%%%%%%%%%%%%%%%%%%%%%%%%%%%%%%%%%%%%%%%%%%%%%%%%%%%%%%%%%%%%%%%%%%%%%%%%%%%%%%%%%%
\section{Word similarity experiments}\label{sec:expe}
We designed a series of experiments for the word similarity task to evaluate how our algorithms compare against other approaches in the literature. In the next sections, we will present and discuss the main characteristics of the training corpus, benchmarks, and compared systems.

\subsection{Training Corpus}\label{ssec:corpus}
We applied our MSSA algorithms to two datasets, to transform their words into synsets using English WordNet 3.0~\citep{Fellbaum:98}, also known as Princeton WordNet, as our lexical database. The datasets are Wikipedia Dumps consisting of wiki articles from April 2010 (WD10)~\citep{Westbury:10} and January 2018 (WD18). Table~\ref{tb:wdump} shows the details for both training corpora after they are cleaned (Section~\ref{ssec:mssa}). %maybe talk about how different the training set are ? Glove/Google/etc all use different datasets for training

\begin{table}[h]
\caption{Dataset token details. WD10 - Wikipedia Dump 2010 (April); WD18 - Wikipedia Dump 2018 (January).} \label{tb:wdump}
\centering
\small
\begin{tabular}{lrrrr}
\toprule
\multicolumn{1}{c}{\multirow{2}{*}{\textbf{POS}}}
& \multicolumn{2}{c}{\textbf{Words ($10^{6}$)}}
& \multicolumn{2}{c}{\textbf{Synsets}}  \\
& \textbf{WD10} & \textbf{WD18}
& \textbf{WD10} & \textbf{WD18}          \\
\midrule
Nouns                                         & 299.41        & 463.31                          & 55731         & 56546                  \\
Verbs                                         & 130.14        & 161.96                          & 11975         & 12237                  \\
Adverbs                                       & 27.25         & 31.17                           & 3091          & 3056                   \\
Adjectives                                    & 75.77         & 104.03                          & 15512         & 15798                  \\
\midrule
\textbf{Total}                                & 532.57        & 760.47                          & 86309         & 87637                  \\
\bottomrule
\end{tabular}
\end{table}

\subsection{Hyperparameters, setup and details}
Once all words in the training corpus are processed into synsets, we use a word2vec implementation to produce our synset embeddings. The hyperparameters are set as follows: CBOW for the training algorithm, window size of 15, minimum word count of 10, hierarchical softmax and vector sizes of 300 and 1000 dimensions. If not specified, all the other hyperparameters were used with their default values\footnote{\url{https://radimrehurek.com/gensim/models/word2vec.html}}. Our system was implemented using Python 3.6, with NLTK 3.2.5 and using the \emph{gensim} 3.4.0~\citep{Rehurek:10} library.

In our experiments, we evaluate our approach with several systems, described in Sections~\ref{ssec:ncws} and~\ref{ssec:cws}, using two different training corpora (WD10 and WD18) for the word similarity task. In a second-level analysis, we also explore the properties of our models separately, over different perspectives. For WD10, we discuss the effects of the number of iterations on our refined model MSSA-NR, with $-N$ ranging from 0 to 2, where $N = 0$ (MSSA) characterizes the initial scenario (Google News vectors) and  $N$$\geq$$1$ characterizes the iterative one (synset vectors), as illustrated in Figure~\ref{fig:Sysarch}. For WD18, we investigate which of our representations of word-senses performs better, the one considering a local context (MSSA) or the global one (MSSA-D). The comparison of our refined model (MSSA-NR) against the MSSA-D algorithm is not explored in the proposed experiments, but we plan to include it in future works. However, to analyze how our synset embeddings are affected by the timestamp difference in the Wikipedia snapshot, we do compare the results of MSSA for both training corpora, WD10 and WD18. The standard number of dimensions used in our experiments is 300, where there is no specific label for MSSA, and 1000, which is indicated with $-T$ next to the algorithm's name.

The differences between metric names, benchmarks, datasets and hyperparameters make it difficult to perform a direct comparison between all available systems. We try to alleviate this situation by explaining the reason behind our choices for the components in our architecture. In the disambiguation step, we use WordNet~\citep{Fellbaum:98} as our lexical database, due to its robustness and popularity for this task. Princeton WordNet (or English WordNet) is the most used resource for WSD in English, also available in more than 70 different idioms~\citep{Navigli:09}. WordNet is also free of charge (for any purpose), can be accessed without any restriction, and is fully integrated with NLTK in Python, making its use preferable over other lexicographic resources, at least for now. As for the word embeddings step, we chose word2vec~\citep{Mikolov_b:13} over other popular techniques, such as GloVe~\citep{Penni:14}, fastText~\citep{Bojanowski:17}, and ELMo~\citep{Matthew:18} because of word2vec's resource-friendly implementation, popularity, and robustness in several NLP tasks~\citep{Iacobacci:15,Iacobacci:16,Li:15,Mancini:17,Taher:16,Rothe:15}. In addition, Glove's embeddings are based on the co-occurrence probabilities of the words in a document, encoded in a word-context co-occurrence matrix (counting) and word2vec's embeddings are built using prediction techniques (CBOW or skip-gram), which are more close to our objective. While GloVe requires the entire matrix to be loaded into memory, making its consumption of RAM quadratic in its input size, the word2vec implementation used in this paper works with linear memory usage, facilitating the training part of our system. As for fastText, its approach uses word substrings ($n$-grams) to produce embeddings, in addition to complete words, as in word2vec. The results comparing word2vec and fastText (no $n$-grams) models are almost equivalent, but some report~\citep{Jain:16} that fastText exhibits better performance in syntactic tasks, in comparison with word2vec, which is more adequate for semantic representations. Since our model is focused on the semantic aspects of each word-sense and WordNet would not be able to provide valid synsets for many of the produced $n$-grams (e.g. kiwi - kiw, iwi), word2vec was a natural choice. Our training corpus is selected so that a greater number of other systems could be compared under the same circumstances. In ELMo, they compute their word vectors as the average of their characters representations, which are obtained through a two-layer bidirectional language model (biLM). This would bring even more granularity to the sub-word embeddings proposed in fastText, as they consider each character in a word will have their own $n$-dimension vector representation. Another factor that prevents us from using ELMo, for now, is its expensive training process\footnote{\url{https://github.com/allenai/bilm-tf}}. We also considered the recently published Universal Sentence Encoder (USE) \citep{Cer:18} from Google, but their implementation does not allow it to be trained in a new corpus such as ours (synset-based), only to use their pre-calculated vectors. WD10~\citep{Westbury:10} is commonly used by many systems~\citep{Chen:15,Huang:12,Iacobacci:15,Li:15,Liu:15,Neela:14} and WD18 is introduced by us as a variation in our experiments to analyze the behavior of our own approaches. %R1/R2 - justifying more some of our choices

Recent publications have pointed out some problems (e.g. model overfitting, subjectivity) in using word similarity tasks to evaluate word embeddings models~\citep{Berg:12,Faruqui:16}. We try to mitigate this situation by illustrating some aspects, such as the general idea of the proposed architecture, a detailed description of the components used in the system, the training corpus specification, hyperparameters' definitions, and comparison of our approaches in different training scenarios. We also apply our models to the most popular benchmarks available, without changing their original structure, and categorize all referenced results according to the correct metrics (AvgSim, AvgSimC, MaxSim, MaxSimC/LocalSim and GlobalSim) defined by seminal authors~\citep{Huang:12,Reisinger:10}. Unfortunately, many authors have not described either what exact metric they used in their experiments or do not specify which one was used by their referenced results. It is common to notice systems being compared under different scopes; this and other issues make our evaluation harder for some specific systems. We try to mitigate such situations by providing as many details as possible for the experiments, metrics, and artifacts used.

The results presented in Sections~\ref{ssec:ncws} and~\ref{ssec:cws} are organized in three blocks for each benchmark (Tables~\ref{tb:rg65}:\ref{tb:scws}), where they are separated by a break line and ordered as follows:

\begin{enumerate}
  \item \textbf{Single-sense embeddings}: traditional word embeddings where all word-senses are collapsed into one vector representation per word. Approaches that concatenate a word vector with their senses are also included;
  \item \textbf{Multi-sense embeddings}: each word-sense has a specific vector representation. Approaches that have a vector for both the word and their senses separately are also included;
  \item \textbf{MSSA embeddings}: all our proposed models, for multi-sense embeddings.
\end{enumerate}

%include interesting characteristics of compared systems

With the exception of MSSG~\citep{Neela:14},~\citep{Chen:14}, and CNN-MSSG~\citep{Chen:15}, which are trained using the skip-gram model, all the compared systems either use CBOW or an independent approach of word embeddings (e.g. GloVe). Results not reported by the authors in their publications are marked as ''-" for the given metrics.

All proposed algorithms (MSSA, MSSA-D, MSSA-NR) and the generated models used in this paper are available in a public repository\footnote{\url{https://github.com/truas/MSSA}}.

\subsection{Benchmark details}\label{ssec:benchmark}
The experiments were separated into two major categories, based on the datasets' characteristics: No Context Word Similarity (NCWS) and Context Word Similarity (CWS). All datasets are widely used in the word similarity task by the compared systems. The former (1 to 5) groups' benchmarks that provide similarity scores for word pairs in isolation, while the latter (6) provides a collection of word pairs with their similarity scores accompanied with sentence examples of their use. These sentences are used to illustrate a context where each word compared is applied. The benchmarks used are described as follows:

\begin{enumerate}
  \item \textbf{RG65}: 65 noun pairs. The similarity scale ranges from 0 to 4~\citep{Rubenstein:65};
  \item \textbf{MC28}: 28 pairs of nouns that were chosen to cover high, intermediate, and low levels of similarity in RG65. This is the same set of words in MC30~\citep{Milles:91}, except for two words not present in WordNet. The similarity scale ranges from 0 to 4~\citep{Resnik:95};
  \item \textbf{WordSim353}: 353 noun pairs divided into two sets of English word pairs, the first set with 153 word pairs and the second with 200~\citep{Finkelstein:02}. The original dataset was later re-organized by~\citep{Agirre:09}, claiming that this dataset did not make any distinction between similarity and relatedness. We used the original version published~\citep{Finkelstein:02}. The similarity scale ranges from 0 to 10;
  \item \textbf{MEN}: 3,000 word pairs, randomly selected from words that occur at least 700 times in the ukWaC and Wacky corpora~\footnote{http://wacky.sslmit.unibo.it/doku.php?id=corpora} combined, and at least 50 times in the \emph{ESP Game}. The similarity scale ranges from 0 to 50~\citep{Bruni:12};
  \item \textbf{SimLex999}: 666 noun-noun pairs, 222 verb-verb pairs, and 111 adjective-adjective pairs. the similarity scale ranges from 0 to 10~\citep{Hill:14};
  \item \textbf{SCWS - Stanford Context Word Similarity}: 2,003 word pairs and their sentential contexts, consisting of 1328 noun-noun pairs, 399 verb-verb pairs, 140 verb-noun, 97 adjective-adjective, 30 noun-adjective, 9 verb-adjective, and 241 same-word pairs. The similarity scale ranges
       from 0 to 10~\citep{Huang:12}.
\end{enumerate}

We tried to keep our basic configuration as close as possible to recent previous publications, so we considered the cosine similarity as our distance measure and report the Spearman correlation value ($\rho$) in our experiments. To guarantee a common scenario between all benchmarks, we normalized their similarity scale to an interval of [-1, 1]. Very few publications reported results for both Spearman and Pearson correlation values, but we considered only the first, to minimize the differences between our comparisons and so that more systems could be included in our paper. The results reported in our experiments, for all model variations, have high significant Spearman order correlation, with a $p-value$ under 0.001, another characteristic that most publications often do not mention. %may

\subsection{No context word similarity}\label{ssec:ncws}
In this section, we evaluate our model against popular approaches available for 5 benchmarks: RG65, MEN, WordSim353, SimLex999 and MC28. We compare our results with:~\citep{Chen:14}, Retro (using Glove with 6 billion words and WordNet with all synsets)~\citep{Faruqui:15},~\citep{Huang:12}, SensEmbed~\citep{Iacobacci:15} (400 dimensions), SW2V (variations using BabelNet and WordNet with UMBC and Wikipedia Dump from 2014)~\citep{Mancini:17}, word2vec (using UMBC and WD14)~\citep{Mancini:17}, word2vec~\citep{Mikolov_b:13}, (NP)MSSG (for 50 and 300 dimensions)~\citep{Neela:14}, Glove (using 6 and 42 billion words)~\citep{Penni:14}, Pruned-TF-IDF~\citep{Reisingerb:10}, and DeConf~\citep{Taher:16}. If not specified, the compared systems use low-dimensional vectors with 300 dimensions each. All of them also use cosine similarity to calculate the distance of words in each benchmark, except SensEmbed, which uses the \emph{Tanimoto} distance~\citep{Tanimoto:57} for their vector comparison. The Tanimoto coefficient is commonly used for binary attributes in vectors, while cosine similarity is applied mainly to non-binary vectors, when their magnitudes are not relevant. In addition, SensEmbed also introduces what they call a \emph{graph vicinity factor}, an argument created to adjust the final similarity score based on the information provided by BabelNet. Even though NASARI~\citep{Camacho:16} achieves impressive results, its comparison with most systems is compromised since they use variations of the traditional benchmarks in their original report. When considering the SimLex999 benchmark, only the noun-noun pairs were evaluated, discarding the other POS (verb-verb and adjective-adjective). Alternative versions for the MC28 and WordSim353 benchmarks were also used, even though the original versions are more common in the literature. For MC28, they considered its earlier version MC30~\citep{Milles:91}, while for WordSim353 they considered the similarity dataset provided by \cite{Agirre:09}. These and other minor aspects would make the comparison against other systems more restrictive and unrealistic. Thus, we decided to leave their results out of our experiments.

%BabelNet as its disambiguation backbone, through Babelfy. 

Tables~\ref{tb:rg65} and~\ref{tb:men} show the results of MSSA against several models for the RG65 and MEN benchmarks, respectively. In both experiments, SensEmbed and DeConf-Sense present the highest results for the AvgSim and MaxSim metrics, followed by one of our models. SensEmbed builds its vectors by using BabelNet as its disambiguation backbone, through Babelfy. BabelNet is composed of several different resources\footnote{https://babelnet.org/about}, including specific lexicons (e.g. GeoNames, Wikiquote, Microsoft Terminology). After the disambiguation step, they train a word2vec model with 400 dimensions. As mentioned above, they introduce what is called a \emph{graph vicinity factor}, a coefficient that combines the structural knowledge from BabelNet's semantic network and the distributional representation of sense embeddings. This factor multiplies AvgSim (Equation~\ref{eq:avgsim}) and GloSim (Equation~\ref{eq:globalsim}) scores by a parameter $\beta$ to re-adjust the similarity measure~\citep{Iacobacci:15}. DeConf-Sense, like our models, relies on less resources to produce its word-sense embeddings. Their approach uses traditional single-sense embeddings (e.g. Google News Vectors) and divides them into separate word-sense vectors according to WordNet's semantic network. They use the Personalized PageRank~\citep{Haveliwala:02} to calculate the semantic relatedness between two synsets in their core~\citep{Taher:16}. All multi-sense embeddings systems surpass single-sense ones, in which, for the GloSim metric, MSSA-2R-T and MSSA(WD10) have the highest results for RG65 and MEN datasets.

%special mark on "Further investigation is necessary to evaluate if MSSA-2R reached its limit or is stuck in a local maximum. Since the disambiguation step is costly for us, at this point, we decided to explore the effects of more iterations in future experiments. If this process is performed in parallel, we can increase and investigate superior values for $N$."

Even though our models did not perform as well as DeConf-Sense for MaxSim, our approach is able to be trained recurrently, improving the quality of its vectors. In RG65, we start with $\rho$ = 0.857 with MSSA(WD10) and move to $\rho$ = 0.872 with MSSA-1R for the MaxSim as shown in Table \ref{tb:rg65}. We tried to increase the number of iterations, but MSSA-2R did not produce better vectors as we expected. Further investigation is necessary to evaluate if MSSA-2R reached its limit or is stuck in a local maximum. Since the disambiguation step is costly for us, at this point, we decided to explore the effects of more iterations in future experiments. If this process is performed in parallel, we can increase and investigate superior values for $N$.  The comparison with SensEmbed is compromised since their model has many differences with the others, including the metric used. However, even using a simpler lexical database (WordNet) our models obtained competitive $\rho$ values for MSSA-1R, MSSA-T, and MSSA-2R-T.

The increase of dimensionality seems to have a positive effect in most word embeddings models, including ours. WD10 and WD18 models presented improvements when each of their models was compared with its 1000-dimensional version, as Tables\ref{tb:rg65} (RG65) and \ref{tb:men} (MEN) show. For WD10, the increase, was on average, 1.61\% for both benchmarks in the total, while for WD18 it was 0.48\% for RG65 and 0.28\% for MEN. Looking only at MSSA, it is hard to affirm that more words would necessarily represent a better result, for if that were true, Glove-42B and Retro-G6B in Table \ref{tb:men} should have more competitive scores, since they were trained over 42 and 6 billion words, respectively. The performance of WD18 for MSSA and MSSA-D is not clear for the global and local contextual aspects, since their results did not improve consistently for all metrics. However, MSSA did obtain better scores for RG65 and MEN.  We were able to fine-tune the hyperparameters of these models in non-reported results, but this just reinforces the findings of \citep{Berg:12,Faruqui:16} with respect to model overfitting for specific benchmarks. We decided to keep our models consistent among all experiments so we could evaluate how well they generalize.

\begin{table}[H]
\caption{Spearman correlation score ($\rho$) on RG65 benchmark. Highest results reported in \textbf{bold} face.} \label{tb:rg65}
\centering
\small
\begin{tabular}{lccc}
\toprule
\multicolumn{1}{c}{\multirow{2}{*}{\textbf{Models}}}        & \multicolumn{1}{c}{\textbf{Avg}}       & \multicolumn{1}{c}{\textbf{Max}} & \multicolumn{1}{c}{\textbf{Glo}}  \\
                       & \multicolumn{1}{c}{\textbf{Sim}}       & \multicolumn{1}{c}{\textbf{Sim}}  & \multicolumn{1}{c}{\textbf{Sim}}   \\
\midrule
GloVe-42B              & -                     & -               & 0.829      \\
GloVe-6B               & -                     & -               & 0.778      \\
\midrule
Retro-G6B              & -                     & -               & 0.767      \\
Retro-G6B-WN           & -                     & -               & 0.842      \\
\midrule
word2vec               & -                     & -               & 0.754      \\
\midrule
\midrule
DeConf-Sense        & -                     & \textbf{0.896}           & -          \\
DeConf-Word           & -                     & 0.761               & -      \\
\midrule
SensEmbed        & \textbf{0.871}                     & 0.894           & -          \\
\midrule
SW2V-Shallow           & -                     & 0.740            & -          \\
SW2V-Babelfy          & -                     & 0.700            & -          \\
\midrule
\midrule
MSSA(WD10)           & 0.779                 & 0.857           & 0.830      \\
MSSA-1R(WD10)          & 0.795                 & 0.872           & 0.825      \\
MSSA-2R(WD10)         & 0.814                 & 0.869           & 0.858      \\
MSSA-T(WD10)           & 0.783                  & 0.878           & 0.845      \\
MSSA-1R-T(WD10)          & 0.825                 & 0.871           & 0.856      \\
MSSA-2R-T(WD10)   & 0.822               & 0.878           & \textbf{0.859}      \\
\midrule
MSSA(WD18)            & 0.828                 & 0.794           & 0.821      \\
MSSA-D(WD18)          & 0.801                 & 0.826           & 0.817      \\
MSSA-T(WD18)      & 0.776                 & 0.847           & 0.816      \\
MSSA-D-T(WD18)     & 0.795                 & 0.839           & 0.835      \\
\bottomrule
\end{tabular}
\end{table}

The results reported by the SW2V algorithm~\citep{Mancini:17} in Table~\ref{tb:men} (MEN), show an interesting behavior with respect to the lexical database used. Their $\rho$ varies by no more than 0.01 when we compare the models using the same corpus (UMBC or WD14), which indicates that the BabelNet (-BN) variation is as robust as WordNet (-WN) to capture the semantic relationships in this dataset. This is supported by our model results as well, since our $\rho$ score fluctuates around the same range with a slightly superior performance for MSSA(WD18), with $\rho$ = 0.769.

\begin{table}[H]
\caption{Spearman correlation score ($\rho$) on MEN benchmark. The results of~\protect\citep{Chen:14} and word2vec were reported by~\protect\citep{Mancini:17}. Highest results reported in \textbf{bold} face. }\label{tb:men}
\centering
\small
\begin{tabular}{lccc}
\toprule
\multicolumn{1}{c}{\multirow{2}{*}{\textbf{Models}}}        & \multicolumn{1}{c}{\textbf{Avg}}       & \multicolumn{1}{c}{\textbf{Max}} & \multicolumn{1}{c}{\textbf{Glo}}  \\
                       & \multicolumn{1}{c}{\textbf{Sim}}       & \multicolumn{1}{c}{\textbf{Sim}}  & \multicolumn{1}{c}{\textbf{Sim}}   \\

\midrule
Retro-G6B                  & -           & -           & 0.737                          \\
Retro-G6B-WN-All           & -           & -           & 0.759                          \\
\midrule
word2vec(UMBC)             & -           & 0.750        & -                              \\
word2vec(WD14)             & -           & 0.720        & -                              \\
\midrule
\midrule
Chen et al.(2014)        & -           & 0.620        & -                              \\
\midrule
DeConf-Sense               & -           & \textbf{0.786}       & -                     \\
DeConf-Word                & -           & 0.732           & -                          \\
\midrule
SensEmbed            & \textbf{0.805}           & 0.779       & -                      \\
\midrule
SW2V-BN-UMBC               & -           & 0.750        & -                              \\
SW2V-WN-UMBC               & -           & 0.760        & -                              \\
SW2V-BN-WD14               & -           & 0.730        & -                              \\
SW2V-WN-WD14               & -           & 0.720        & -                              \\
\midrule
\midrule
MSSA(WD10)                & 0.751        & 0.745        & 0.760                      \\
MSSA-1R(WD10)              & 0.781        & 0.751        & 0.790                    \\
MSSA-2R(WD10)             & 0.777         & 0.737       & 0.788                    \\
MSSA-T(WD10)           & 0.778                  & 0.753           & 0.785      \\
MSSA-1R-T(WD10)          & 0.783                 & 0.747           & 0.791      \\
MSSA-2R-T(WD10)   & 0.785                 & 0.744           & \textbf{0.795}      \\
\midrule
MSSA(WD18)                & 0.745         & 0.769       & 0.775                    \\
MSSA-D(WD18)              & 0.768         & 0.716       & 0.765                    \\
MSSA-T(WD18)      & 0.769                 & 0.749           & 0.776      \\
MSSA-D-T(WD18)     & 0.772                 & 0.717           & 0.767      \\
\bottomrule
\end{tabular}
\end{table}

In Table~\ref{tb:wsim353} (WordSim353), all results perform worse than the single-sense embeddings of GloVe~\citep{Penni:14} for the GloSim metric. However, to reach this score they processed 42 billion tokens, while, when considering just 6 billion tokens,  its performance decreases 13.30\%. We, on the other hand, with a little less than 540 million tokens for WD10 can obtain superior results with MSSA-2R and MSSA-2R-T. Even though Pruned-TF-IDF~\citep{Reisingerb:10} follows with a competitive $\rho$ score, their model does not use low-dimensions for their vectors, which makes its direct comparison problematic. In addition, their model relies on several parameter adjustments (e.g. pruning cutoff, feature weighting, number of prototypes,feature representation). Our model works independently of any parameters other than those required by the word2vec implementation.

For the MaxSim metric we noticed that our initial model MSSA(WD18) obtained equal results when compared to SensEmbed, and better results when considering SW2V in their two forms, -Shallow and -Babelfy. Because of our model's simplicity we highlighted MSSA(WD18) instead of SensEmbed for this metric. These are all models that use BabelNet in their disambiguation process, while ours only uses WordNet, which BabelNet incorporates completely. In addition, most of our models for WD10 and WD18 also present superior scores against SW2V-Babelfy.

Under more similar characteristics, MSSG and NP-MSSG models of~\citep{Neela:14}, present results much more closer to our systems for both AvgSim and GloSim, as Table~\ref{tb:wsim353} shows. They also produce multi-sense embeddings, based on word-senses, which are learnt jointly with the word vector itself. MSSG and NP-MSSG only differ in the number of senses a word can have, which is similar to what we accomplish with MSSA. Their training time of 6 hours for MSSG-300d and 5 hours for NP-MSSG are comparable with our synset embeddings (Section~\ref{ssec:s2v}) step. However, unlike MSSA, which requires a disambiguation process prior to the embeddings one, their model does these tasks at the same time and with strong competitive results. For GloSim, MSSG-300d and NP-MSSG-300d present $\rho$ of 0.692 and 0.691 respectively, while MSSA-T(WD18) and MSSA-D-T has values of 0.692 and 0.693, respectively. This shows that, as for the amount of words processed, the number of dimensions will not necessarily provide better results. In general, our models using 300 dimensions had a better performance than those with a higher dimensionality. The same behavior is observed when we consider the values using AvgSim for MSSG-300d, MSSA-1R-T, and MSSA-D-T. In most cases, the increase of dimensionality and iterations in our models showed a slightly negative impact in the overall score for WD10 and WD18, with respect to their 300 and 1000 dimension models.

As explained in Section~\ref{ssec:benchmark}, the WordSim353 benchmark is composed of two separate datasets, the first with 153 word pairs and the second with 200. According to \cite{Agirre:09}, this benchmark conflates two distinct aspects of linguistics: similarity and relatedness, so they released separate datasets for each characteristic. Thus, some authors~\citep{Iacobacci:15,Mancini:17} take this into account and use the updated version. In Table 4 of~\citep{Iacobacci:15}, they report their results for the WordSim353 dataset without any distinction or reference to the structure suggested in~\citep{Agirre:09}, so we assumed they were making use of the original version proposed in~\citep{Finkelstein:02}. We were hoping that our model corresponding to the global (MSSA-D) context would be able to handle such nuances, but results showed otherwise. If we only analyze the performance of both MSSA(WD18) and MSSA-D, we notice the inconsistency of the results, with the former approach showing better results than the latter for MaxSim and GloSim. It is in our plans to investigate which subcategory can be better explored by our models for this benchmark, but for now we kept WordSim353 as one single set so more systems could be compared.

%The results reported in~\cite{Iacobacci:15}, for this benchmark, considered the sub-divisions suggested in~\cite{Agirre:09}, in which WordSim353 is divided in WS-Sim and WS-Rel. However, they also reported their results for wordsim353 but with no specification if this was considering the original or the one from 2009 agirre

\begin{table}[H]
\caption{Spearman correlation score ($\rho$) on WordSim353 benchmark.~\protect\citep{Huang:12} results were reported by~\protect\citep{Neela:14}. Highest results reported in \textbf{bold} face.} \label{tb:wsim353}
\centering
\small
\begin{tabular}{lccc}
\toprule
\multicolumn{1}{c}{\multirow{2}{*}{\textbf{Models}}}       & \multicolumn{1}{c}{\textbf{Avg}}       & \multicolumn{1}{c}{\textbf{Max}} & \multicolumn{1}{c}{\textbf{Glo}}  \\
                       & \multicolumn{1}{c}{\textbf{Sim}}       & \multicolumn{1}{c}{\textbf{Sim}}  & \multicolumn{1}{c}{\textbf{Sim}}   \\
\midrule
GloVe-42B            & -      & -      & \textbf{0.759}      \\
GloVe-6B             & -      & -      & 0.658      \\
\midrule
Retro-G6B            & -      & -      & 0.605      \\
Retro-G6B-WN-All     & -      & -      & 0.612      \\
\midrule
\midrule
Huang et al. (2012) & 0.642  &        & 0.228      \\
\midrule
MSSG-50d             & 0.642  & -      & 0.606      \\
MSSG-300d            & 0.709  & -      & 0.692      \\
NP-MSSG-50d          & 0.624  & -      & 0.615      \\
NP-MSSG-300d         & 0.686  & -      & 0.691      \\
\midrule
Pruned-TF-IDF        & -      & -      & 0.734      \\
\midrule
SensEmbed      & \textbf{0.779}     & 0.714  & -          \\
\midrule
SW2V-Shallow         & -      & 0.710  & -          \\
SW2V-Babelfy         & -      & 0.630  & -          \\
\midrule
\midrule
MSSA(WD10)           & 0.725  & 0.702  & 0.727      \\
MSSA-1R(WD10)         & 0.711  & 0.661  & 0.712      \\
MSSA-2R(WD10)        & 0.730  & 0.662  &\textbf{ 0.737}      \\
MSSA-T(WD10)           & 0.712   & 0.669       & 0.721      \\
MSSA-1R-T(WD10)          & 0.708 & 0.666       & 0.716      \\
MSSA-2R-T(WD10)   & 0.729        & 0.667       & \textbf{0.737}      \\
\midrule
MSSA(WD18)           & 0.663  & \textbf{0.714}  & 0.712      \\
MSSA-D(WD18)         & 0.708  & 0.626  & 0.702      \\
MSSA-T(WD18)      & 0.694                 & 0.637           & 0.692      \\
MSSA-D-T(WD18)     & 0.702                 & 0.623           & 0.693      \\
\bottomrule
\end{tabular}
\end{table}

\begin{table}[H]
\caption{Spearman correlation score ($\rho$) on SimLex999 benchmark.~\protect\citep{Chen:14} and word2wec results were reported by~\protect\citep{Mancini:17}. Highest results reported in \textbf{bold} face.} \label{tb:simlex999}
\centering
\small
\begin{tabular}{lccc}
\toprule
\multicolumn{1}{c}{\multirow{2}{*}{\textbf{Models}}}       & \multicolumn{1}{c}{\textbf{Avg}}       & \multicolumn{1}{c}{\textbf{Max}} & \multicolumn{1}{c}{\textbf{Glo}}  \\
                       & \multicolumn{1}{c}{\textbf{Sim}}       & \multicolumn{1}{c}{\textbf{Sim}}  & \multicolumn{1}{c}{\textbf{Sim}}   \\
\midrule
word2vec(UMBC)             & -      & 0.390  & -                              \\
word2vec(WD14)            & -      & 0.380  & -                              \\
\midrule
\midrule
Chen et al.(2014)       & -      & 0.430  & -                              \\
\midrule
DeConf-Sense               & -      & \textbf{0.517}  & -                              \\
DeConf-Word                & -      & 0.443     & -                          \\
\midrule
SW2V-BN-UMBC               & -      & 0.470  & -                              \\
SW2V-WN-UMBC               & -      & 0.450  & -                              \\
SW2V-BN-WD14               & -      & 0.430  & -                              \\
SW2V-WN-WD14               & -      & 0.430  & -                              \\
\midrule
\midrule
MSSA(WD10)                 & 0.427  & 0.368  & 0.396                          \\
MSSA-1R(WD10)               & 0.438  & 0.369  & 0.405                          \\
MSSA-2R(WD10)              & 0.440  & 0.369  & 0.408                         \\
MSSA-T(WD10)            & 0.456   & 0.393     & 0.432     \\
MSSA-1R-T(WD10)          & 0.468  & 0.394     & \textbf{0.441}      \\
MSSA-2R-T(WD10)         & \textbf{0.469}   & 0.385     & 0.439      \\
\midrule
MSSA(WD18)                 & 0.375  & 0.438  & 0.404                          \\
MSSA-D(WD18)               & 0.401  & 0.351  & 0.374                          \\
MSSA-T(WD18)      & 0.460                 & 0.389           & 0.430      \\
MSSA-D-T(WD18)     & 0.425                 & 0.372           & 0.391     \\
\bottomrule
\end{tabular}
\end{table}

The last two NCWS benchmarks, SimLex999 and MC28, are particularly challenging for distinct reasons. For SimLex999 in Table~\ref{tb:simlex999}, we noticed a consistent improvement with respect to the increase in dimensionality between models of the same configuration using WD10, but not for WD18. Results using the refined models and their 1000-dimensional versions also presented consistent improvement for WD10. However, overall our models performed poorly regardless of their configuration for all metrics, while DeConf-Sense presented the best results for MaxSim. The average Spearman correlation values for this dataset seems to be low in all publications, rarely surpassing $\rho=0.50$. Even in our unreported models, we do not have satisfactory results. The same behavior was observed when we tried to apply our model in no-nouns benchmarks, such as: YP130~\citep{Yang:06} and SimVerb3500~\citep{Gerz:16}. For the former, our Spearman scores were on average $\rho=0.563$, while for the latter, $\rho=0.243$ (MaxSim). Our suspicion is that our models, as with most compared systems, are not robust enough to deal with datasets of this nature (no nouns). It seems verbs, adjectives, and adverbs-based benchmarks need a more focused approach to deal with their characteristics properly. As in Table~\ref{tb:men} (MEN), the results of the SW2V algorithm~\citep{Mancini:17} for the MaxSim metric presented little or no variation in their $\rho$ score, when considering the same training corpus under different lexical databases, as Table~\ref{tb:simlex999} (SimLex999) shows. Likewise, our MSSA trained in the WD18 corpus obtained the best result among our models.

For MC28, reported in Table~\ref{tb:mc28}, the lack of recent publications makes it hard to draw any strong conclusions about compared models. We do obtain the same results as SensEmbed, but with a much simpler architecture and less resources, since all of our algorithms only use WordNet as its lexical resource. If we consider ACL State-of-the-art Wiki\footnote{\url{https://aclweb.org/aclwiki/MC-28_Test_Collection_(State_of_the_art)}} we would have obtained the third best result considering the human upper bound as the gold standard. Since MC28 is a subset of RG65, our models presented similar results, but with slightly better results on average for both the WD10 and WD18 training corpora.

%We suspect most authors choose to focus on RG65, given that MC28 is a subset of this.

\begin{table}[H]
\caption{Spearman correlation score on MC28 benchmark. Highest results reported in \textbf{bold} face.} \label{tb:mc28}
\centering
\small
\begin{tabular}{lccc}
\toprule
\multicolumn{1}{c}{\multirow{2}{*}{\textbf{Models}}}       & \multicolumn{1}{c}{\textbf{Avg}}       & \multicolumn{1}{c}{\textbf{Max}} & \multicolumn{1}{c}{\textbf{Glo}}  \\
                       & \multicolumn{1}{c}{\textbf{Sim}}       & \multicolumn{1}{c}{\textbf{Sim}}  & \multicolumn{1}{c}{\textbf{Sim}}   \\
\midrule
GloVe-42B                  & -      & -      & 0.836                          \\
GloVe-6B                   & -      & -      & 0.727                          \\
\midrule
\midrule
SenseEmbed                & -      & 0.880      & -                          \\
\midrule
\midrule
MSSA(WD10)                 & 0.833  & 0.862  & 0.842                          \\
MSSA-1R(WD10)               & 0.825  & 0.883 & 0.843                          \\
MSSA-2R(WD10)              & 0.829  & 0.849  & 0.847                          \\
MSSA-T(WD10)            & \textbf{0.845}      & \textbf{0.888}           & \textbf{0.875}      \\
MSSA-1R-T(WD10)          & 0.841     & 0.883         & 0.862      \\
MSSA-2R-T(WD10)         & 0.801      & 0.866           & 0.836      \\
\midrule
MSSA(WD18)                 & 0.775  & 0.799  & 0.792                          \\
MSSA-D(WD18)               & 0.835  & 0.807  & 0.829                          \\
MSSA-T(WD18)      & 0.796                & 0.834           & 0.818      \\
MSSA-D-T(WD18)     & 0.801                 & 0.833           & 0.821      \\
\bottomrule
\end{tabular}
\end{table}

In all benchmarks, except SimLex999, our proposed models give competitive results within the top three positions. Considering the MaxSim and GloSim metrics, we report either the first or second highest Spearman correlation values in all experiments. Our models are strongly based on the context surrounding the word we disambiguate, so we believe, in theory, that the more information about a word's surroundings we have, the more accurate our representations will be. Nonetheless, our models performed very well when applied to benchmarks considering the similarity of word pairs without any extra context information. In addition, we were able to obtain better results than several systems based on more complex architectures and on lexical databases. MSSA's single training phase minimizes the hyperparameter adjustments to the word2vec implementation only, which makes it easier to replicate. Moreover, our modular configuration displays an appealing layout for us to apply the components that have obtained better results throughout the experiments for the word similarity task. Considering the NCWS category, MSSA showed superior results over MSSA-D, making us suspect that a local context sliding window seems to be more adequate to extract semantic features of a corpus to build their embeddings. As in other compared systems, we also noticed that the increase of dimensionality is most likely to improve the overall performance for the word similarity task, but this is not always true. The increase of dimensionality usually adds more computational time to the word embeddings step, which might not be worth the effort, given the small differences when compared to lower-dimensional vector models.

\subsection{Context word similarity}\label{ssec:cws}
In this section the results reported by AutoExtend (Synsets)~\citep{Rothe:15} and CNN-VMSSG~\citep{Chen:15} were also incorporated. For the SCWS benchmark, we did not report the results for the MaxSim metric, since almost all publications do not report them, as well. As explained in Section~\ref{ssec:benchmark}, the SCWS dataset provides word pairs, their similarity score, and a given specific context for each word in a sentence, alleviating dubious interpretations~\citep{Huang:12}.

Table~\ref{tb:scws} shows DeConf~\citep{Taher:16}, NP-MSSG-300d~\citep{Neela:14} and MSSA(WD10) with the highest Spearman correlation values, in descending order for AvgSim. However, when considering MaxSimC and GloSim, the results of MSSA-1R and MSSA(WD10) give state-of-the-art scores, respectively. It seems that the extra information about a word's context indeed helps our model to better select a word-sense, but using all the remaining metrics for the word similarity task did not produce good results. Since in our disambiguation step MSSA looks for a word-sense with the highest similarity of its word ($w_{i}$) against its neighbors ($w_{i-1}$ and $w_{i+1}$), the score obtained in a metric that reflects the maximum similarity of a word-sense, given its sentence context, is expected. 

This encourages us to apply our models to tasks where sentences are compared, instead of just words. A natural direction to our work it to extend our techniques to document classification problems, where each entity (document) will have a collection of tokens to build their embeddings. CNN-VMSSG presents the second highest results for MaxSimC and GloSim after our models, but their approach relies on two training steps, one for the convolutional neural network (CNN) part and one for the multi-sense skip-gram (MSSG), instead of just one training step, like ours.
%#R1-4

The increase in the word-vector dimensionality had the same behavior as in previous experiments, with the exception of MSSA(WD18) which had an increase of almost 8\% on its score for AvgSim if compared with MSSA-D. The refined model (MSSA-NR) applied in the WD10 corpus and the context approach (MSSA-D) for WD18 seemed to have little or no improvement in the overall score when compared with their initial models (MSSA).

For AvgSimC, our models did not present competitive results, while DeConf and MSSG-300d were able to produce top scores. Since our approach is oriented towards the higher similarity between word-senses and their context, perhaps a different scheme to select prospective word-senses could improve our system. It would also be interesting to apply the top-ranked algorithms to our model and compare their performance for all metrics. Most published results do not report their findings for all metrics defined in~\citep{Huang:12,Reisinger:10}, making their direct comparison arduous. Unfortunately, DeConf is designed to use pre-trained single dimensional vectors to produce their multi-sense embeddings, so our approach would not be easily applicable, as we produce a vector for each word-sense directly. NP-MSSG~\citep{Neela:14} and SW2V~\citep{Mancini:17} on the other hand, offer the necessary flexibility to use our annotated corpus to produce new embeddings.

\begin{table}[H]
\caption{Spearman correlation score ($\rho$) on SCWS benchmark.~\protect\citep{Huang:12} results were reported by~\protect\citep{Neela:14}. Highest results reported in \textbf{bold} face.} \label{tb:scws}
\centering
\small
\begin{tabular}{lcccc}
\toprule
\multicolumn{1}{c}{\multirow{2}{*}{\textbf{Models}}}        & \multicolumn{1}{c}{\textbf{Avg}}       & \multicolumn{1}{c}{\textbf{Avg}} & \multicolumn{1}{c}{\textbf{Max}} & \multicolumn{1}{c}{\textbf{Glo}} \\
                       & \multicolumn{1}{c}{\textbf{Sim}}       & \multicolumn{1}{c}{\textbf{SimC}}  & \multicolumn{1}{c}{\textbf{SimC}} & \multicolumn{1}{c}{\textbf{Sim}}  \\
\midrule
GloVe-42B              & -      & -       & -       & 0.596      \\
GloVe-6B               & -      & -       & -       & 0.539      \\
\midrule
\midrule
AutoExtend     & 0.626  & 0.637   & -       & -          \\
\midrule
Chen et al. (2014)    & 0.662  & 0.689   & -       & 0.642      \\
\midrule
CNN-VMSSG          & 0.657  & 0.664   & 0.611   & 0.663      \\
\midrule
DeConf-Sense           & \textbf{0.708}  & \textbf{0.715}   & -       & -            \\
\midrule
Huang et al. (2012)  & 0.628  & 0.657   & 0.261   & 0.586     \\
\midrule
MSSG-50d               & 0.642  & 0.669   & 0.492   & 0.621      \\
MSSG-300d              & 0.672  & 0.693   & 0.573   & 0.653      \\
NP-MSSG-50d            & 0.640  & 0.661   & 0.503   & 0.623      \\
NP-MSSG-300d           & 0.673  & 0.691   & 0.598   & 0.655      \\
\midrule
Pruned-TF-IDF          & 0.604  & 0.605   & -       & 0.625      \\
\midrule
SensEmbed        & -      & 0.624       & 0.589   & -          \\
\midrule
\midrule
MSSA(WD10)             & 0.667  & 0.581   & 0.637   & \textbf{0.667}      \\
MSSA-1R(WD10)           & 0.660  & 0.581   & \textbf{0.639}   & 0.659      \\
MSSA-2R(WD10)          & 0.665  & 0.593   & 0.631   & 0.665      \\
MSSA-T(WD10)           & 0.659  & 0.590   & 0.617    & 0.664      \\
MSSA-1R-T(WD10)        & 0.655  & 0.594   & 0.623   & 0.658      \\
MSSA-2R-T(WD10)        & 0.661  & 0.604   & 0.617   & 0.664      \\
\midrule
MSSA(WD18)             & 0.593  & 0.569   & \textbf{0.639}   & 0.651      \\
MSSA-D(WD18)           & 0.640  & 0.557   & 0.613   & 0.640      \\
MSSA-T(WD18)           & 0.649  & 0.588   & 0.617   & 0.654      \\
MSSA-D-T(WD18)         & 0.638  & 0.570   & 0.597   & 0.639      \\
\bottomrule
\end{tabular}
\end{table}

%%%%%%%%%%%%%%%%%%%%%%%%%%%%%%%%%%%%%%%%%%%%%%%%%%%%%%%%%%%%%%%%%%%%%%%%%%%%%%%%%%%%%%%%%%%%%%%%%%%%%%%%%%%%%%%%%%%%%%%%%%
\section{Further Discussions and Limitations}\label{sec:limitations} %R2-3,4,5,6
In this section, we try to provide a deeper discussion about the main aspects of our techniques by pointing out their strengths and limitations, while discussing alternatives that can be taken into account.

The main objective of our algorithms is to properly transform word-based documents into synset-based ones, that will be used in systems or tasks that deal with semantic representation at some level. For this, we use WordNet to identify possible word-senses of a given word. Unfortunately, this forces us only to work with formal texts (i.e. free of colloquial English, slang and typos). Traditional word embeddings techniques are derived directly from the raw text, what can be either an advantage or disadvantage, depending on which task is selected for its validation. If we were exploring a document classification task in which the documents were based on informal texts (e.g. user comments in a blog, movies reviews) our approaches would probably work poorly. On the other hand, we believe if the documents considered were using formal English (e.g. scientific paper abstracts) our techniques would present good results. 

Another aspect is the WordNet structure itself, with respect to the number of its available synsets and its idiom-version. Currently, in version 3.0., WordNet has 155,287 unique strings mapped, with 117,659 synsets, which leaves a reasonable amount of words out if we consider the entire English vocabulary. An alternative lexical database to be considered is BabelNet~\citep{Navigli:12}, which is composed of several different resources\footnote{https://babelnet.org/about} (including WordNet) and specific lexicons (e.g. GeoNames\footnote{http://verbs.colorado.edu/~mpalmer/projects/verbnet.html}, Wikiquote\footnote{https://www.wikiquote.org}, Microsoft Terminology\footnote{http://www.microsoft.com/Language/en-US/Terminology.aspx}). Since we are using the English version of WordNet (also known as Princeton WordNet) our system currently does not apply to other languages in its current version. The language aspect is not a barrier to BabelNet since it is a multilingual lexical database. However, it is important to mention that there are other idiom versions\footnote{\url{http://globalwordnet.org/wordnets-in-the-world/}} of WordNet available \citep{WNPri:10} that can be incorporated in our techniques.

From the experiments presented in Sections \ref{ssec:ncws} and \ref{ssec:cws} we noticed that our technique is very sensitive to the presence of verbs and adjective POS, as illustrated by Table~\ref{tb:simlex999}. Even though we did not explicitly report the results obtained using YP130~\citep{Yang:06} and SimVerb3500~\citep{Gerz:16} datasets, our overall performance was not satisfactory. However, when using datasets that provide context to the words being evaluated (SCWS), our techniques showed competitive results (Table~\ref{tb:scws}). This encourages us to explore tasks in which we can use a larger context to support our decision, such as document classification, sentiment analysis, and plagiarism detection.

Our modular and flexible architecture (Figure~\ref{fig:Sysarch}), provides an interesting setup, that can be applied to any expert system using natural text as its input. This is because the disambiguation and annotation steps work as a pre-processing phase, and can be applied to the raw text directly in order to obtain a more precise semantic representation for a given word. Nevertheless, since MSSA considers all word-senses available in WordNet, it might not be the best option for very large training sets. In our experiments, it took us approximately four full days to transform all documents in the English Wikipedia Dump of 2010 \citep{Westbury:10}. However, if available, one can parallelize the processing of the words into synsets since our MSSA, MSSA-NR and MSSA-D are performed on a document level individually. Moreover, the word to synset transformation task only needs to be performed once. After the annotated synset training corpus is performed, one can use it in any desired activity (e.g. NLP tasks, training a word embeddings model). In addition, since words that do not exist on WordNet will be automatically discarded, this will reduce the time of any word embeddings technique applied to the translated training corpus. As with the experiments in Section \ref{sec:expe}, this can lead to good and bad results depending on the datasets used for validation.   

Differing from MSSA, which evaluates all senses for each word and requires the glosses vectors calculated, the MSSA-NR algorithm only considers the word-senses that are actually embbeded in the first place. Therefore, the non-used word-senses are dropped, reducing the amount of comparisons required for each context window. In this case, the average complexity for MSSA-NR is smaller than the one in MSSA. Approaches like MSSG~\citep{Neela:14} fix the number of possible word-senses available and obtain a faster disambiguation process than MSSA. The gain in speed however, comes with the price of removing word-senses that could provide a better semantic representation for words.

During our experiments, we also found that MSSA-D is highly affected by the number of words in a document, probably due to its global search. Dijkstra's algorithm does perform a blind search for all the available paths in the graph that goes from the word-senses of the first word to the last. MSSA and MSSA-NR, on the other hand, have a local search approach and deal with the word-senses inside each context window, one at a time. Thus their processing time is, at some level, dependent on the number of word-senses in the context.

In our pipeline, we train a synset vector representation model from scratch, so we do not take advantage of pre-trained models (e.g. Google News, GloVe). Approaches like \citep{Taher:16,Rothe:15} follow this direction and do save some time in the disambiguation task. However, they are not able to produce their own vector representation directly from the training corpus. In addition, their approach requires some parameter tuning, adding a certain complexity to the system as a whole. In contrast, our technique can be applied directly to any training corpus, resulting in a new representation that can be transferred to several different problems. For example, we can apply MSSA to better represent sentences provided to chatbots in support systems and improve the quality of their answers. Additionally, MSSA, MSSA-NR, and MSSA-D are completely unsupervised, so they do not rely on any parameters other than those required by the word embeddings algorithm. 

Lastly, another important aspect explored by our approach is the re-usability of MSSA-NR. To the best of our knowledge, the ability to iteratively use the produced synset vectors to improve the word sense disambiguation task and provide a more refined synset annotation is not explored by any of the compared systems we found. This opens many new directions on how word embeddings and word sense disambiguation can mutually benefit each other.    

%%%%%%%%%%%%%%%%%%%%%%%%%%%%%%%%%%%%%%%%%%%%%%%%%%%%%%%%%%%%%%%%%%%%%%%%%%%%%%%%%%%%%%%%%%%%%%%%%%%%%%%%%%%%%%%%%%%%%%%%%%
\section{Final considerations}\label{sec:concl}
In this paper, we proposed an algorithm called MSSA that automatically disambiguates and annotates any text corpus using a sliding context window for each word. We have confirmed that single vector representation limitations can be mitigated by applying MSSA to a traditional word2vec implementation, producing more robust multi-sense embeddings with minimum hyperparameter tuning. Additionally, we performed an extensive comparison with many recent publications in the word similarity task and categorized their results according to standard metrics (Section~\ref{sec:mse}). We noticed some shortcomings in several publications with respect to the setup of their experiments, particularly considering the metrics available in their experiments. Most publications focus on one or two metrics at the same time, ignoring the others. Hence, it is hard to confirm the superiority of one system over another. However, we tried to mitigate this by comparing systems with the greatest number of similarities we could find.

%convergence issue
We showed that the combination between the proposed MSSA algorithms and word2vec is able to give us solid results in 6 different benchmarks: RG65, MEN, WordSim353, SimLex999, MC28 and SCWS. In our refined model (MSSA-NR), we explored how we can build and improve the produced synset embeddings model iteratively. Word similarity is a downstream task and somewhat independent of whether the produced word-sense embeddings converge or not. Therefore, we believe specific experiments to study the values of our embeddings are still necessary to fully understand their behavior. The other group of our models, using global (MSSA-D) and local (MSSA) context information are also used to build synset embeddings. The former approach finds the most similar word-senses from the first word to the last in a document in the disambiguation step, while the second approach looks for the most suitable synset given a defined sliding window. Initially, we thought that MSSA-D would produce the best result on average, since it considers the whole document as its global context. However, if we analyze the results of WD18 only, this is not true. Most of our experiments showed that MSSA obtained better results when compared to MSSA-D. Apparently, features of the local context window are more powerful than those obtained globally. We also noticed that a dimensional increase in the model proved more effective than a change in the approach itself, but at the cost of some extra computation time. The simplicity  of our model makes its use attractive, while the independent components of its architecture allows its extension to other NLP tasks beyond word similarity.

%#R2-7
Since all proposed MSSA techniques are performed in the raw text directly and prior to any word embeddings training model, they can easily be incorporated into any NLP pipeline, independent from the problem or task. This opens new alternatives to different real-world problems or systems that make use of natural language text as their input. In particular, some would directly benefit from a better semantic representation, especially in the expert systems arena. In monitoring social media activity, one could use MSSA to improve the quality of the processed comments about a certain company or product and evaluate its digital reputation (e.g. customer surveys). In chat-bots, by helping intelligent systems to comprehend human textual interaction could lead us to a more human-like perception in services over multiple scenarios (e.g. tutoring systems, automated health system, technical support). In the same direction, virtual digital assistants that are able to differentiate the nuances in a human discourse can definitely provide a better service with respect to the need and characteristics required by the user. Another interesting option, closer to academia, would be to explore the semantic signature between authors in scientific papers and principal investigators (PI's) in research grants. The correlation between scientific papers and research grants awarded would help us to identify more relevant features that lead authors to a high productivity in their area. In the recommender systems arena, more specifically in scientific paper recommendation, one could use our semantic annotation to explore more characteristics in articles other than title, abstract, and keywords. The inner sections of a paper (e.g. introduction, related work, conclusions) are rich in content and would much benefit from our algorithms. In general, any system that requires more semantic features in order to support our decision making process can benefit from a the proposed techniques in this paper.

%#R2-7  #R1-4
Considering how the proposed algorithms help to better represent multi-sense words in the word similarity task, a natural question is how they would perform in other NLP downstream tasks. Basically, the MSSA algorithms can be applied in different tasks with minimum effort, as long as these tasks use natural text as their input. In document classification, we would be dealing with the similarity of a collection of words from a common class. In theory, MSSA would benefit from such a scenario since each word collection would be under the same semantic structure (i.e. document subject). In the text summarization problem, one could use the dispersion of word-senses to find the most relevant semantic topics, using them to create a topic model representation based on their semantic signature. We plan to move forward in these areas, and use MSSA and its variants in many interesting downstream tasks.    

Currently, our model considers a sliding context window of +/- 1 tokens, unigrams or non-stemmed words, but we intend to pursue some extensions, such as keeping common n-grams, having a flexible context sliding window size for MSSA and different weighting schemes for the context analysis. We will pursue a weighting scheme whereby context words closer to the target word-sense have more importance. In addition, we plan to evaluate higher levels of our refined model (MSSA-NR). This seems to be a more certain path to follow than just increasing the dimensionality for each scenario. In addition we want to explore new alternatives to build semantic representations using MSSA as their base. Finally, we also would like to integrate MSSA with the best-ranked systems evaluated throughout our experiments.

%%%%%%%%%%%%%%%%%%%%%%%%%%%%%%%%%%%%%%%%%%%%%%%%%%%%%%%%%%%%%%%%%%%%%%%%%%%%%%%%%%%%%%%%%%%%%%%%%%%%%%%%%%%%%%%%%%%%%%%%%%
\section{Acknowledgements}
This work was partially supported by the Science Without Borders Brazilian Government Scholarship Program, CNPq  [grant number 205581/2014-5];

%%%%%%%%%%%%%%%%%%%%%%%%%%%%%%%%%%%%%%%%%%%%%%%%%%%%%%%%%%%%%%%%%%%%%%%%%%%%%%%%%%%%%%%%%%%%%%%%%%%%%%%%%%%%%%%%%%%%%%%%%%

\bibliographystyle{unsrtnat}
\bibliography{main.bib}

\end{document}